\documentclass[journal]{IEEEtran}

\usepackage[utf8]{inputenc}
\usepackage[T1]{fontenc}
\usepackage{hyperref}
\usepackage{url}
\usepackage{booktabs}
\usepackage{amsfonts}
\usepackage{nicefrac}
\usepackage{microtype}
\usepackage{xcolor}
\usepackage{amsmath}
\usepackage{amssymb}
\usepackage{amsthm}
\usepackage{graphicx}
\usepackage{subcaption}
\usepackage{algorithm}
\usepackage{algpseudocode}
\usepackage{tikz}
\usetikzlibrary{arrows.meta,positioning,shapes.geometric,calc,fit,backgrounds}
\usepackage{enumitem}
\usepackage{multirow}
\usepackage{cite} % sorts and compresses numeric citation lists, e.g. [2]--[4]

\newtheorem{definition}{Definition}


\definecolor{sgInk}{HTML}{263238}
\definecolor{sgBlue}{HTML}{3366A6}
\definecolor{sgPurple}{HTML}{7451A6}
\definecolor{sgTeal}{HTML}{168C88}
\definecolor{sgOrange}{HTML}{D97732}
\definecolor{sgGreen}{HTML}{3A8F5B}
\definecolor{sgRed}{HTML}{B55252}
\definecolor{sgGold}{HTML}{C89520}
\definecolor{sgSoftGray}{HTML}{F4F6F8}

\newcommand{\sys}{\textsc{EpiCoach}}
\newcommand{\EIG}{\mathrm{EIG}}

\newcommand{\ACWR}{\mathrm{ACWR}}
\newcommand{\dd}{\mathrm{d}}
\newcommand{\E}{\mathbb{E}}

\newcommand{\calD}{\mathcal{D}}
\newcommand{\calH}{\mathcal{H}}

\newcommand{\HR}{\mathrm{HR}}
\newcommand{\VP}{\mathrm{VP}}
\newcommand{\dodo}[1]{\mathrm{do}(#1)}

\newcommand{\NRides}{223}
\newcommand{\NActivities}{653}
\newcommand{\NDays}{556}
\newcommand{\NSegments}{24}
\newcommand{\NSegmentsTen}{15}
\newcommand{\NTraversals}{458}
\newcommand{\NHours}{296}
\newcommand{\DateStart}{2 April 2025}
\newcommand{\DateEnd}{30 September 2026}
\newcommand{\Span}{eighteen months}

\title{The Athlete as a Scientist of One:\\
Agentic Bayesian Experimental Design for Individualized Physiology\\
from Consumer Wearables}

\author{\IEEEauthorblockN{Behnam Moradi}\\
\IEEEauthorblockA{Software Department, KavAI\\California, USA\\behnam.moradi@kavai.com}}

\begin{document}
\maketitle

\begin{abstract}
Large language model (LLM) agents for personal health read wearable data and recommend, but do not \emph{learn about the individual} in a way that can be audited or falsified. We present \sys{}, an agentic system that treats one athlete, a cross-country mountain biker, as the subject of accountable N-of-1 science. Claims about the athlete live in an evidence-gated \emph{hypothesis registry}: each must issue a probabilistic prediction about a future session \emph{before} it occurs, and its belief moves only through computed evidence from a logged statistical fit. Predictions come from a mechanistic digital twin inferred from heart rate, GPS and barometric altitude: a lagged heart-rate response model, a fitness--fatigue model with learned time constants, and a latent technical-descending skill state, measured on discovered trail segments. The agent prescribes the next training week as a constrained experiment maximizing information gain subject to load and safety constraints, around two instrument rides. We evaluate the system two ways over eighteen months of data. A rolling-origin replay benchmark shows the twin beats standard forecasters on heart rate but not segment time, and the agentic loop matches a static model. A live prospective phase runs the weekly cycle: in its first three weeks the ledger sealed 20 predictions, scored 14, the descent forecast calibrated after one correction, the climb forecast pessimistic as the athlete improved, and the safety officer vetoed the agent twice. Every correction is preserved in the registry, a template for auditable, falsifiable personalization in digital health. We release the registry, benchmark and protocol.
\end{abstract}


%%%%%%%%%%%%%%%%%%%%%%%%%%%%%%%%%%%%%%%%%%%%%%%%
\section{Introduction}
\label{sec:intro}

A coach who has worked with the same athlete for years knows things about that athlete that no textbook contains: that her heart rate climbs by six beats per hour once the temperature passes $28^{\circ}$C, that two short nights cost her more on technical descents than on climbs, that she holds fitness longer than the forty-two-day constant assumed by every training platform. This knowledge is individual, it was acquired by watching many sessions, and, crucially, it was \emph{tested}: the coach predicted, observed, and revised. Consumer wearables now record the raw material for this kind of knowledge continuously, and LLM agents can reason over it. Yet the systems built so far do something quite different from what the experienced coach does.

Recent personal-health agents are impressive readers of wearable data. PH-LLM fine-tunes Gemini to produce expert-rated coaching from thirty-day summaries~\cite{cosentino2025phllm}; PHIA answers open-ended numerical questions over wearable time series by generating and running code~\cite{merrill2026phia}; the Personal Health Agent orchestrates data-scientist, domain-expert and coach sub-agents~\cite{heydari2025pha}; SePA grounds advice in personalized risk models and trusted retrieval~\cite{ozolcer2025sepa}; LLM-SPTRec generates training plans from a sports-science knowledge graph~\cite{he2026sptrec}. All of them share a direction of information flow: data in, recommendation out, population knowledge applied to an individual. None maintains a body of \emph{claims about this particular person} that is explicit, probabilistic, revised by evidence, and exposed to falsification. If such a system tells the athlete that her heart rate drifts in the heat, nothing records whether that claim was ever tested, what it would take to overturn it, or whether it was quietly abandoned.

We argue that the missing object is not a better recommender but an \emph{epistemic loop}, and that building one for a single human raises problems that automated-science systems in chemistry or biology~\cite{lu2024aiscientist,ghareeb2025robin,takahara2026hep} do not face. The subject cannot be reset; every observation costs a training session that also changes the system being studied; the measurements are noisy wrist sensors rather than assays; and the designer has a second objective, making the athlete better, that competes with learning about her. We call the resulting problem \emph{epistemic coaching} and make four design commitments:

\begin{enumerate}[leftmargin=*,itemsep=1pt]
\item \textbf{A prediction ledger, not a chat log.} Every hypothesis in the registry must emit a probabilistic prediction about a specific future ride, timestamped before the ride. Beliefs move only through a tool that requires a logged statistical fit. This is what makes $N{=}1$ science defensible: it closes the garden of forking paths~\cite{gelman2013garden,nosek2018preregistration} that would otherwise let a fluent language model explain any outcome after the fact.
\item \textbf{A mechanistic twin from a watch alone.} We infer, from optical heart rate, GPS and barometric altitude only, a lagged heart-rate response with drift and heat interactions, a fitness--fatigue model whose time constants are \emph{estimated} rather than assumed, and a latent technical-skill state that is deliberately separated from aerobic fitness because heart rate stops measuring mechanical work on descents~\cite{gregory2007relationship,impellizzeri2007physiology}.
\item \textbf{Workouts as experiments, with safety as a hard constraint.} The next ride is chosen from a menu by a utility that trades expected information gain against expected performance improvement, subject to acute:chronic workload bounds, heart-rate caps and readiness gates enforced in code rather than by the language model. Repeated trail segments, discovered automatically, serve as the measurement instrument.
\item \textbf{Evaluation by prediction, not by plausibility.} We score the system by rolling-origin replay over \Span{} of history and by a live prospective phase in which every prediction is sealed before the ride it concerns, using proper scoring rules and forecast-comparison tests designed for a single long time series.
\end{enumerate}

\paragraph{Contributions.}
(i)~A formalization of coaching as sequential, safety-constrained Bayesian experimental design over an evidence-gated hypothesis registry with a pre-registered prediction ledger (Section~\ref{sec:problem}). (ii)~A wrist-only digital twin for mountain biking with explicit identifiability analysis of its fitness--fatigue component (Section~\ref{sec:twin}). (iii)~An agent architecture, implemented in Google ADK, whose protocol, not its prompts, guarantees that LLMs propose, select and explain while deterministic tools fit, score and gate (Section~\ref{sec:agent}). (iv)~A public rolling-origin replay benchmark for individualized physiological forecasting with retrospective results on \Span{} of data, and the first three weeks of a live, pre-registered prospective $N{=}1$ phase with its complete audit trail (Section~\ref{sec:eval}).

%%%%%%%%%%%%%%%%%%%%%%%%%%%%%%%%%%%%%%%%%%%%%%%%
\section{Related Work}
\label{sec:related}

\paragraph{LLMs and agents over wearable data.}
PH-LLM~\cite{cosentino2025phllm} fine-tunes a frontier model on expert case studies built from thirty-day aggregates and reasons in-model. PHIA~\cite{merrill2026phia} is a ReAct agent~\cite{yao2023react} with a Python runtime and web search, evaluated on thousands of insight questions. The Personal Health Agent~\cite{heydari2025pha} adds orchestration across specialist sub-agents; PhysioLLM~\cite{fang2024physiollm} adds a statistical-correlation layer; SePA~\cite{ozolcer2025sepa} adds personalized predictive models and whitelisted retrieval. These systems answer questions and give advice. None keeps a persistent, probabilistic, falsifiable model of the user, none chooses what data to collect next, and none writes an intervention back to the device. We use PHIA-style agents as a baseline and PHA-style orchestration as a point of contrast.

\paragraph{AI scientists and hypothesis management.}
End-to-end automated research systems~\cite{lu2024aiscientist,ghareeb2025robin} close a hypothesize--experiment--analyze loop in silico or in the wet lab, where replicates are cheap and the system under study is stationary. The Hypothesis Evolution Protocol~\cite{takahara2026hep} externalizes hypotheses as registry objects with evidence-gated beliefs, lineage and lifecycle; we adopt this discipline and add the ledger requirement that every hypothesis issues a dated prediction about an unobserved outcome. Box's loop with language models~\cite{li2024automated,box1976science} and the Automatic Statistician~\cite{lloyd2014automatic} use LLMs to propose and criticize probabilistic programs on fixed data; BoxingGym~\cite{gandhi2025boxinggym} benchmarks LLM agents on information-gain-driven design in simulated environments and finds them weak at it. Our setting is BoxingGym's real-world counterpart with a living subject, irreversible state and a dual objective.

\paragraph{Training-load modelling and digital twins.}
The Banister impulse-response model~\cite{banister1975systems,banister1991modeling} and the Busso variable-dose variant~\cite{busso2003variable} relate training load to performance through fitness and fatigue filters; they are known to be ill-conditioned~\cite{hellard2006assessing,vermeire2022fitness} and sensitive to the load metric~\cite{vermeire2021influence}. Kalman-filter variants track parameters online~\cite{kolossa2017performance}; optimal-control and reinforcement-learning methods optimize load schedules \emph{given} fitted parameters. We treat parameter uncertainty as the object to be reduced by design, and we use only wrist-derived load and segment-based performance rather than power meters or laboratory tests.

\paragraph{N-of-1 trials and self-experimentation.}
Within-person crossover trials have a mature design and analysis literature~\cite{kravitz2014design,duan2022personalized,daza2024causal,daza2022motr}. CASEbot~\cite{ishita2026casebot} uses an LLM to help users \emph{design} such experiments. In our system the agent generates candidate experiments from model residuals, selects them by expected information gain, executes them through the athlete's device, and analyses them with randomization-respecting estimators.

\paragraph{Mountain-bike physiology.}
Cross-country mountain biking is characterized by sustained high heart rates that remain elevated on descents despite low power output, attributed to isometric load and arousal~\cite{gregory2007relationship,impellizzeri2007physiology}; performance is multidimensional and course-specific~\cite{novak2018multidimensional}; rolling resistance varies two- to three-fold with surface~\cite{bertucci2013mountain}. These facts shape the twin: heart rate is a fitness signal only on climbs, descending is a separate skill, and rolling resistance is a segment-level nuisance.

%%%%%%%%%%%%%%%%%%%%%%%%%%%%%%%%%%%%%%%%%%%%%%%%
\section{Coaching as Accountable Sequential Experimentation}
\label{sec:problem}

\subsection{Setting}
An athlete produces a stream of rides $r=1,2,\dots$ and daily wellness records. Each ride yields a time series $x_r=\{(t,\HR_t,\mathbf{g}_t,a_t,T_t)\}$ of heart rate, GPS position, barometric altitude and temperature, plus device-level summaries. Let $\theta\in\Theta$ denote the parameters of a generative model of the athlete (Section~\ref{sec:twin}) and $p_t(\theta)$ the posterior given all data observed up to time $t$. A \emph{workout} $d\in\calD$ is a specification the athlete can execute (archetype, duration, heart-rate target, segment, day), and $y\sim p(y\mid\theta,d)$ is the vector of outcomes the twin predicts for it.

\subsection{Hypothesis registry and prediction ledger}
\begin{definition}[Hypothesis]
A hypothesis $H=(\text{text},\phi_H,\pi_H)$ consists of a natural-language claim; a formal predicate $\phi_H:\Theta\to\{0,1\}$ or a sub-model $M_H$; a belief $\pi_H\in(0,1)$; together with a lifecycle state in \{\textsc{proposed}, \textsc{under\_test}, \textsc{supported}, \textsc{refuted}, \textsc{undetermined}, \textsc{retired}\} and parent pointers recording refinement and merging.
\end{definition}

Two rules distinguish our registry from a free-form notebook. \textbf{Evidence gating} (inherited from \cite{takahara2026hep}): $\pi_H$ changes only through \texttt{record\_evidence}($H$, $\log\mathrm{BF}$, \texttt{fit\_id}), where $\log\mathrm{BF}$ is a log Bayes factor or an equivalent evidence score computed by a deterministic fit whose identifier exists in the statistician's store, and the update is $\operatorname{logit}\pi_H\leftarrow\operatorname{logit}\pi_H+\log\mathrm{BF}$~\cite{kass1995bayes}. The language model has no tool that sets $\pi_H$ directly. \textbf{Ledger requirement} (ours): a hypothesis may enter \textsc{under\_test} only after $\texttt{predict}(H,\ r^{\ast},\ q_H)$ has registered a predictive distribution $q_H$ for a specified future ride $r^{\ast}$, and the registry refuses the call if its timestamp is later than the ride's start. Ledger predictions are scored with proper scoring rules~\cite{gneiting2007strictly} when $r^{\ast}$ is observed, and the scores are the only admissible evidence for hypotheses about predictive claims. \textbf{Identifiability gate} (ours): for a directional claim about a parameter, a Savage--Dickey-style $\log\mathrm{BF}$ is admitted only if the posterior standard deviation has shrunk by at least $10\%$ relative to the prior; otherwise the evidence is recorded as zero (the test still consumes budget). Without this rule a bounded prior that is merely re-shaped by weak data can masquerade as evidence; it did so for the fitness time constant in our own data (Section~\ref{sec:ffm}). The verdict \textsc{undetermined} is reachable: a hypothesis whose evidence has not left $|\log\mathrm{BF}|<\log 3$ after a pre-set budget of tests is closed as unresolved rather than left open indefinitely.

\subsection{The design problem}
At each decision point the agent chooses the next workout. Let $\pi=(\pi_H)_{H\in\calH}$ be current registry beliefs and $\theta$ the twin parameters. The expected information gain of a design about $\theta$ is the mutual information~\cite{lindley1956measure,chaloner1995bayesian}
\begin{equation}
\begin{split}
\EIG_\theta(d)&=\E_{p_t(\theta)\,p(y\mid\theta,d)}\!\left[\log p(y\mid\theta,d)-\log p(y\mid d)\right],\\
p(y\mid d)&=\int p(y\mid\theta,d)\,p_t(\theta)\,\dd\theta,
\end{split}
\label{eq:eig}
\end{equation}
and the analogous quantity for the discrete hypothesis set is the expected reduction in registry entropy, $\EIG_\calH(d)=\mathbb{H}(\pi)-\E_{y\mid d}[\mathbb{H}(\pi\mid y,d)]$, computed by Bayesian model averaging over the sub-models $M_H$. Because the athlete is not a laboratory, we do not maximize information alone. With $\Delta p(d)$ denoting the twin's predicted change in performance over a fixed horizon if $d$ is executed, the agent solves
\begin{equation}
\begin{split}
d^{\star}_t=\arg\max_{d\in\calD_t^{\mathrm{feas}}}\;&\lambda_t\,\widehat{\EIG}(d)\\
&+(1-\lambda_t)\,\E_{p_t}[\Delta p(d)],
\end{split}
\label{eq:utility}
\end{equation}
where $\lambda_t\in[0,1]$ decays over a training block (explore early, exploit late) and $\calD_t^{\mathrm{feas}}\subset\calD$ is the set of designs that satisfy hard constraints: acute:chronic workload ratio $\ACWR\in[0.8,1.3]$ computed on 7:28-day rolling loads~\cite{gabbett2016training}, weekly load within $\pm15\%$ of the block plan, at least one rest day per week, a heart-rate cap, no prescription when device-reported training readiness falls below a threshold or the athlete reports illness, and calendar availability. Constraints are evaluated by code; the language model may argue for a design but cannot admit an infeasible one. We call the resulting policy \emph{information-directed training}.

%%%%%%%%%%%%%%%%%%%%%%%%%%%%%%%%%%%%%%%%%%%%%%%%
\section{A Wrist-Only Mechanistic Twin for Mountain Biking}
\label{sec:twin}

The twin is deliberately minimal: each component exists because a hypothesis class needs it, and each parameter has a physiological reading. All models are fit by Hamiltonian Monte Carlo~\cite{hoffman2014nuts} in NumPyro~\cite{phan2019composable}; cheap Kalman-form updates run weekly and full refits monthly. Appendix~\ref{app:models} gives priors and full specifications.

\subsection{Signal quality and derived quantities}
Optical wrist heart rate fails in characteristic ways: it locks onto pedalling cadence or arm-shake frequency, drops out on rough descents, and lags true heart rate by $20$--$45$\,s~\cite{bent2020investigating,navalta2020concurrent}. We compute a per-sample quality flag (cadence-lock when $|\HR_t-\text{cadence}_t|<4$\,bpm for $\geq30$\,s with collapsed variance; implausible jumps $|\Delta\HR|>8$\,bpm/s), use a Student-$t$ likelihood on heart rate, and expose the fraction of good samples as a covariate so that instrument failure is itself something the registry can reason about. Altitude is taken from the barometer, never from GPS. The device used here records no cadence, temperature or power in its FIT records, so cadence lock is detected indirectly through its flatline-and-release signature; ambient temperature, humidity and conditions come from the activity's weather record (reported in Fahrenheit by the vendor API and converted).

Without a power meter we recover \emph{virtual power} on climbs from the standard road-cycling power balance~\cite{martin1998validation},
\begin{equation}
\begin{split}
\VP_t=\Big[&m g v_t(\sin\theta_t+C_{rr}\cos\theta_t)\\
&+\tfrac12\rho C_dA\,v_t^3+m a_t v_t\Big]/\eta,
\end{split}
\label{eq:vp}
\end{equation}
with total mass $m$, grade $\theta_t$, speed $v_t$ and acceleration $a_t$. On climbs of grade $\geq4\%$ below $15$\,km/h the gravitational term dominates and the aerodynamic term is negligible; the rolling coefficient $C_{rr}$, which varies two- to three-fold with surface~\cite{bertucci2013mountain}, is absorbed as a segment-level random effect. Virtual power is used only on climb segments and only as an \emph{input} to the heart-rate response; it is never reported as an estimate of mechanical power.

\subsection{Repeated segments as instruments}
\label{sec:segments}
Repeated terrain converts an uncontrolled ride into a repeated-measures design. We build the athlete's trail network from OpenStreetMap: all named ways tagged as paths, tracks or service roads within each riding area are fetched, ways sharing a name are joined end-to-end into a \emph{chain}, and each chain is densified to $5$\,m points carrying an along-trail coordinate $s$. Every GPS sample is snapped to the nearest chain point within $20$\,m, which gives it a trail name and a scalar position; a \emph{run} is a contiguous stretch of samples on one chain with monotone $s$, and its sign gives the direction (the same trail ridden up and down is two segments). Because riders enter and leave a trail at different points, timing the whole trail would inflate variance. We therefore define a segment by \emph{consensus gates}: the longest stretch of $s$ covered by at least $60\%$ of the rides that touch the chain (and by at least five rides). A traversal is a pass from the entry gate to the exit gate with monotone progress and no stop longer than $90$\,s; entry and exit times are interpolated at the gate crossings. Each traversal records moving time (the primary timing outcome; wall-clock time is inflated by trail-side stops), heart-rate summary and quality, mean virtual power, ambient temperature, elapsed ride time, recent precipitation and gear. Appendix~\ref{app:segments} gives thresholds and failure modes; Appendix~\ref{app:data} lists the discovered segments.

\subsection{Heart-rate response on climbs}
\label{sec:hrmodel}
For climb traversals we model the steady-state heart rate the body is approaching and the lag with which it approaches it:
\begin{align}
\HR_{t+1}&=\HR_t+\frac{\Delta t}{\tau_{hr}}\big(\HR^{ss}_t-\HR_t\big)+\varepsilon_t,\notag\\
&\qquad \varepsilon_t\sim t_\nu(0,\sigma_{hr}),\label{eq:hrlag}\\
\HR^{ss}_t&=\alpha_{r}+\beta_{w(r)}\,\VP_t+\gamma_T\,(T_r-20)+\gamma_d\,\tfrac{t}{3600}\notag\\
&+\gamma_{dT}(T_r-20)\tfrac{t}{3600}+u_{s(t)}.\label{eq:hrss}
\end{align}
Here $\alpha_r$ is a ride-level intercept (day-to-day arousal and resting heart rate), $u_s$ a segment random effect that absorbs $C_{rr}$ and surface, $\gamma_T$ the heat offset, $\gamma_d$ the cardiac-drift slope in bpm per hour~\cite{coyle2001cardiovascular}, and $\gamma_{dT}$ its interaction with temperature. The slope $\beta_w$ follows a Gaussian random walk over weeks $w$, so that a contrast $\beta_{w_2}-\beta_{w_1}$ is the formal content of the hypothesis ``my climbing response has improved''. The pair $(\gamma_d,\gamma_{dT})$ is the formal content of ``my heart rate drifts $X$\,bpm/h above $Y^{\circ}$C''. A segment-aggregate version regresses traversal-mean heart rate on traversal-mean virtual power with the same covariates and fits in seconds; the sample-level recursion is reserved for the final analysis.

\paragraph{Retrospective fit.}
On $188$ climb traversals over $9$ segments the aggregate model (NUTS, no divergences, $\hat R\leq1.003$) gives $\beta=19\pm3$\,bpm per $50$\,W, cardiac drift $\gamma_d=6.0$\,bpm/h ($90\%$ interval $2.1$--$9.6$), a heat offset $\gamma_T=0.34\pm0.23$\,bpm/$^{\circ}$C and a heat--drift interaction indistinguishable from zero ($\gamma_{dT}=-0.27\pm0.38$; only $17$ rides exceeded $25^{\circ}$C). Heart rate at reference power changed by $+4\pm4$\,bpm from the first to the last week, so the claim ``aerobic response improved'' has posterior probability $0.16$. Residual scale is $6.9$\,bpm with $R^2=0.65$.

\subsection{Technical descending as a latent skill}
On descents heart rate decouples from mechanical work~\cite{gregory2007relationship}, so we do not model it as a fitness signal. Instead, descent time at matched conditions measures a latent skill:
\begin{equation}
\begin{split}
\log T_{r,s}&=\mu_s+\lambda_s\,S_{w(r)}+\delta_{\mathrm{wet}}\,\mathrm{wet}_r\notag\\
&\quad+\delta_{\mathrm{light}}\,\mathrm{light}_r+\boldsymbol{\delta}^{\top}\mathbf{c}_r+\eta_{r,s},\\
S_w&\sim\mathcal{N}(S_{w-1},\sigma_S^2),
\end{split}
\label{eq:skill}
\end{equation}
where $\lambda_s$ is a segment loading (technical segments load more on skill), $\mathbf{c}_r$ holds candidate covariates such as prior-night sleep duration and elapsed ride time, and $S_w$ is a weekly random walk. ``Sleep below $6.5$\,h slows my technical descents by $Y\%$'' is a coefficient in $\boldsymbol{\delta}$ and is testable only on segments with sufficient $\lambda_s$. Descent heart rate is retained as an exploratory ``arousal cost'' outcome. On $104$ stop-free descents over $8$ segments the fit gives $\delta_{\mathrm{sleep}}=-0.022\pm0.015$ per hour, i.e.\ one hour less sleep slows a descent by $2.2\%$ ($90\%$ interval $-0.2$ to $+4.7\%$); wet conditions show no effect; the log-timing residual is $7.9\%$. The ledger prediction for the instrument descent is drawn from this posterior: segment level, skill state advanced one random-walk step, and covariates at protocol conditions (dry, habitual sleep, median elapsed ride time). A ``stop-free'' traversal is one whose clock starts when the rider rolls \emph{out} of the entry gate, so a pause at the gate (lap press, regroup) does not disqualify it; before this rule only $5$ of $25$ instrument descents entered the fit.

\begin{figure*}[t]
\centering
\resizebox{\linewidth}{!}{%
\begin{tikzpicture}[
  font=\sffamily\scriptsize,
  box/.style={draw=sgInk!55,fill=sgSoftGray,rounded corners=1.2mm,line width=0.6pt,minimum height=9mm,text width=23mm,align=center,inner sep=2pt},
  llm/.style={box,draw=sgPurple,fill=sgPurple!8},
  tool/.style={box,draw=sgTeal,fill=sgTeal!8},
  gate/.style={box,draw=sgRed!80!sgInk,fill=sgRed!7},
  store/.style={box,draw=sgGold,fill=sgGold!12,text width=26mm},
  world/.style={box,draw=sgBlue,fill=sgBlue!8,text width=20mm},
  flow/.style={-{Stealth[length=1.8mm]},draw=sgInk!80,line width=0.65pt},
  pflow/.style={flow,draw=sgPurple},
  dflow/.style={flow,dashed,draw=sgBlue},
  lab/.style={font=\sffamily\tiny,text=sgInk!75,fill=white,inner sep=1pt}
]
% ---- top row: observe -> model -> hypothesize
\node[world] (garmin) {\textbf{Garmin Connect}\\rides, wellness};
\node[tool,right=5mm of garmin] (ingest) {\textbf{Ingest}\\sync, HR QC, segments};
\node[tool,right=5mm of ingest] (stat) {\textbf{Statistician}\\fit twin, residuals, LOO};
\node[llm,right=5mm of stat] (hyp) {\textbf{Hypothesizer}\\anomalies $\to$ claims};
\node[llm,right=5mm of hyp] (mod) {\textbf{Modeler} $\leftrightarrow$ \textbf{Critic}\\template spec, $\leq 3$ rounds};
% ---- middle row: stores
\node[store,below=11mm of stat,xshift=-6mm] (fits) {\textbf{Fit store}\\posteriors, \texttt{fit\_id}};
\node[store,right=5mm of fits] (reg) {\textbf{Hypothesis Registry}\\beliefs, lineage, ledger};
\node[llm,right=5mm of reg] (scribe) {\textbf{Scribe}\\weekly report to athlete};
% ---- bottom row: design -> gate -> device (right to left)
\node[llm,below=27mm of mod] (des) {\textbf{Experiment Designer}\\menu $\calD_t$, constraints};
\node[tool,left=5mm of des] (eig) {\textbf{EIG scorer}\\nested MC, Eq.~\eqref{eq:utility}};
\node[gate,left=5mm of eig] (safe) {\textbf{Safety Officer}\\ACWR, HR cap, readiness};
\node[tool,left=5mm of safe] (push) {\textbf{Workout Pusher}\\JSON $\to$ calendar};
\node[world] (athlete) at (garmin |- push) {\textbf{Athlete + watch}};
% ---- top row flows
\draw[flow] (garmin) -- (ingest);
\draw[flow] (ingest) -- (stat);
\draw[flow] (stat) -- node[lab,above]{residuals} (hyp);
\draw[flow] (hyp) -- node[lab,above]{claims} (mod);
\draw[flow] (mod.north) -- ++(0,3mm) -| node[lab,pos=0.25,above]{refit candidate sub-models} (stat.north);
% ---- into stores
\draw[flow] (stat.south) -- node[lab,left]{posteriors} (fits.north);
\draw[flow] (stat.south) -- ++(0,-4mm) -| node[lab,pos=0.3,below]{\texttt{record\_evidence}} ([xshift=-7mm]reg.north);
\draw[pflow] (hyp.south) -- node[lab,right]{\texttt{register}, \texttt{predict}} ([xshift=7mm]reg.north);
\draw[flow] (fits) -- (reg);
\draw[flow] (reg) -- (scribe);
\draw[flow] (mod.south) -- (des.north);
% ---- bottom row flows
\draw[flow] (des) -- node[lab,above]{$\calD_t$} (eig);
\draw[flow] (eig) -- node[lab,above]{ranked} (safe);
\draw[flow] (safe) -- node[lab,above]{$d^\star$} (push);
\draw[flow] (push) -- (athlete);
\draw[dflow] (athlete) -- node[lab,left]{rides, wellness} (garmin);
\draw[flow] (reg.south) -- ++(0,-3mm) -| node[lab,pos=0.25,below]{$\pi,\;p_t(\theta)$} ([xshift=-6mm]eig.north);
\end{tikzpicture}}
\caption{\footnotesize The weekly \sys{} cycle. Purple nodes are LLM agents; teal nodes are deterministic tools; the red node is the safety gate; gold nodes are persistent stores. LLMs may \texttt{register} hypotheses and \texttt{predict}, but only the Statistician's \texttt{record\_evidence}, keyed by a stored \texttt{fit\_id}, can move a belief. The Safety Officer's constraints are code and cannot be argued around.}
\label{fig:system}
\end{figure*}
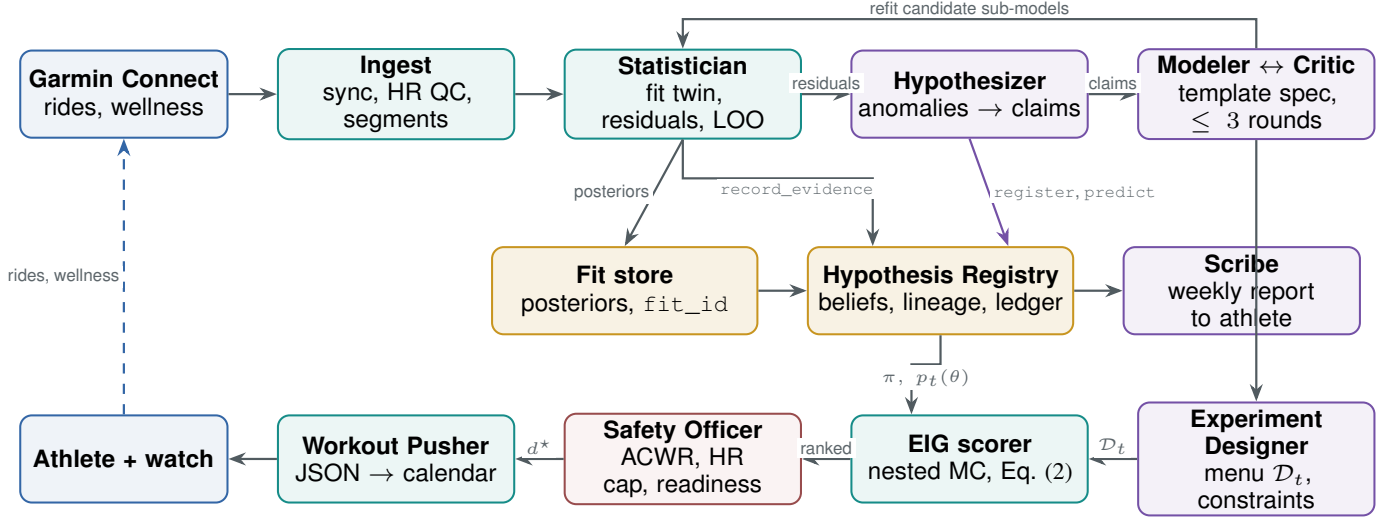

\subsection{Fitness--fatigue with learned time constants}
\label{sec:ffm}
Daily load $w_d$ is computed from heart rate alone, using Banister's exponential TRIMP~\cite{banister1991modeling} as the primary metric and Edwards zone-TRIMP~\cite{edwards1993heart} and the device's proprietary load as sensitivity inputs, since load metrics are known not to be interchangeable~\cite{vermeire2021influence}. Fitness and fatigue are exponential filters of load,
\begin{equation}
\begin{split}
F_d&=e^{-1/\tau_1}F_{d-1}+w_d,\qquad G_d=e^{-1/\tau_2}G_{d-1}+w_d,\\
p_d&=p_0+k_1F_d-k_2G_d,
\end{split}
\label{eq:ffm}
\end{equation}
with the Busso variable-gain extension~\cite{busso2003variable} available when posterior checks demand it. Because there is no laboratory test, performance $p_d$ is measured as heart-rate-adjusted speed on climb segments (log speed regressed on segment, heart rate, temperature, elapsed time and $p_d$, i.e.\ the external output at matched internal cost), with the device's reported VO$_2$max available as an optional noisy auxiliary channel. Priors are informative but wide, $\tau_1\sim\mathrm{LogNormal}(\ln 42,0.3)$, $\tau_2\sim\mathrm{LogNormal}(\ln 7,0.4)$, $k_2/k_1\sim\mathrm{LogNormal}(\ln 2,0.3)$.

\paragraph{Identifiability is a result, not a nuisance.}
The fitness--fatigue model is ill-conditioned even with weekly laboratory tests~\cite{hellard2006assessing,vermeire2022fitness}: $\tau_1,\tau_2,k_1,k_2$ trade off along ridges of the likelihood. Our data reproduce this. With weakly informative log-normal priors the posterior escapes to $\tau_1\approx180$ days and $\tau_2\approx145$ days: a slow trend, not a training response. Bounding $\tau_1\in[15,90]$, requiring $\tau_2<\tau_1$ and $k_2>k_1$, and adding an annual harmonic for trail conditions returns a posterior that is essentially the prior: Bhattacharyya overlap $0.89$ for $\tau_1$, $0.95$ for $\tau_2$, $0.99$ for $k_2/k_1$; the $90\%$ interval of $k_1$ includes zero; the implied performance swing over \Span{} is $1.9\%$; and leave-one-out comparison cannot separate the full model from a fitness-only model ($|\Delta\mathrm{elpd}|<4$). Without the seasonal term the model absorbs a $14\%$ seasonal swing into ``fitness''; load and season are confounded because the athlete rides more in summer. A synthetic check that simulates speeds from a true $\tau_1=35$ days on the athlete's \emph{actual} load history and observation schedule recovers the latent performance trajectory well ($r=0.95$--$0.99$) but returns $\tau_1$ intervals of roughly $[28,65]$ days, biased upward (Appendix~\ref{app:synthetic}). The observational design, not the model, is the bottleneck.

Three consequences follow. First, the registry reports ``$\tau_1<42$ days'' as unresolved: the identifiability gate (Section~\ref{sec:problem}) refused the raw $\log\mathrm{BF}$ of $+1.45$ because the posterior had \emph{widened} relative to the prior. Second, this is precisely where information-directed design earns its keep: the design that most reduces uncertainty about $\tau_2$ is a planned deload followed by a standardized test climb, and about $\tau_1$ a sustained block; a conventional schedule may never produce either contrast. Third, no real posterior is interpreted before its recovery has been verified on synthetic athletes (Section~\ref{sec:synthetic}).

%%%%%%%%%%%%%%%%%%%%%%%%%%%%%%%%%%%%%%%%%%%%%%%%
\section{The \sys{} Agent}
\label{sec:agent}

\sys{} is a weekly cycle implemented as a Google ADK \texttt{SequentialAgent} with one inner \texttt{LoopAgent}~\cite{googleadk2025} (Figure~\ref{fig:system}). The architecture is not novel for being multi-agent; orchestrated specialist agents already exist for wearables~\cite{heydari2025pha}. What we claim is the \emph{allocation}: every operation that produces a number the system will act on is performed by a deterministic tool, and the language model is confined to the operations where its strengths are real and its failure modes are survivable.

\paragraph{What the LLM does.}
The \textbf{Hypothesizer} receives structured residual diagnostics from the Statistician (e.g.\ ``heart-rate residual $+6$\,bpm on rides with ambient temperature $>27^{\circ}$C, $n{=}14$''; ``descent times on segment~7 bimodal since week~61''), retrieves candidate mechanisms from the literature, and proposes hypotheses with a formal predicate, a prior log-odds and a test plan. This is mechanism retrieval and analogical reasoning, a genuine language-model capability. The \textbf{Modeler} translates an accepted hypothesis into a model specification drawn from a unit-tested template library (add covariate, add interaction, add changepoint, add random walk, split a parameter by condition); it cannot emit free-form model code, which keeps it from inventing unidentifiable structures. A \textbf{Critic} reviews the specification against identifiability red flags and registry lineage, and the pair iterate at most three times. The \textbf{Experiment Designer} proposes the candidate workout menu $\calD_t$ (20--40 designs: long aerobic rides, climb repeats on segment $s$ at heart-rate target $h$, descent laps, deloads, a standardized eight-minute test climb; crossed with duration, day and time of day), negotiates calendar and wellness constraints with the athlete in natural language, and explains the final choice. The \textbf{Scribe} writes the weekly report, including counterfactuals such as ``had you slept $7$\,h, the twin predicts climb~3 would have been $4\%$ faster'' (computed by the twin as a posterior predictive under $\dodo{\text{sleep}=7}$ and merely narrated by the model).

\paragraph{What tools do.}
\textbf{Ingest} synchronizes rides and wellness, runs quality control and segment matching. The \textbf{Statistician} fits the twin and candidate sub-models, computes residual reports, leave-one-out comparisons~\cite{vehtari2017practical} and Bayes factors, and stores every fit under a \texttt{fit\_id}. The \textbf{EIG scorer} evaluates Eq.~\eqref{eq:utility} over $\calD_t$ by nested Monte Carlo~\cite{rainforth2018nesting} with $N{=}M{=}500$ posterior draws (variational estimators~\cite{foster2019variational} are a drop-in if cost demands). The \textbf{Safety Officer} evaluates the feasibility set $\calD_t^{\mathrm{feas}}$ and vetoes; an LLM explains the veto to the athlete but cannot lift it. The \textbf{Workout Pusher} serializes $d^\star$ as a cycling workout with heart-rate-target steps and writes it to the athlete's calendar, from which the watch receives it on the next sync (Appendix~\ref{app:push}).

\paragraph{Information-directed design in practice.}
Scoring the menu against the retrospective posterior ranks an overload week followed by full rest with three repeated test efforts highest ($0.17$ nats), ahead of a three-week block-and-taper ($0.16$) and all conventional weeks ($\leq0.06$): the contrast in which fatigue dissipates while fitness persists is what separates $\tau_2$ from $\tau_1$, as anticipated in Section~\ref{sec:ffm}. Absolute gains per cycle are small because the posterior on $k_1$ is near zero; information accrues over cycles. In a live run on the athlete's current state (acute load $1.26\times$ chronic) the Safety Officer vetoed ten of the top-ranked plans for ACWR, and the Designer settled on a back-loaded conventional week with one test: an instance of safety constraints overriding information gain by construction.

\begin{figure}[!t]
\centering
\includegraphics[width=\columnwidth]{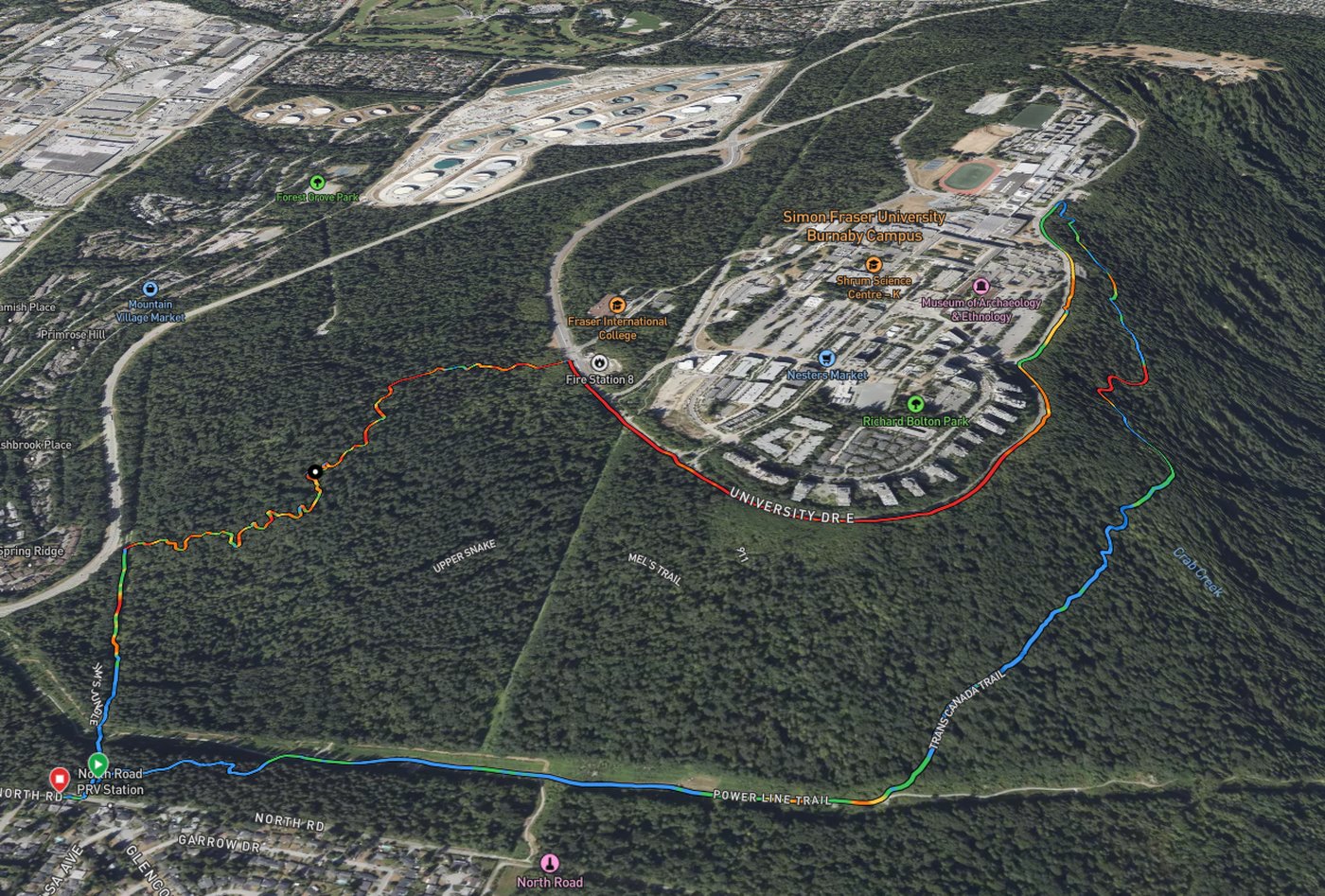}
\caption{\footnotesize The instrument ride on Burnaby Mountain (one recording, coloured by heart-rate zone; satellite view from the vendor's app). From the car park at lower left the athlete warms up along the Power Line Trail, climbs the Trans Canada Trail (a wide, steady-gradient gravel track under canopy) to University Drive, traverses the campus ring road, and descends Nicoles Trail, a narrow technical singletrack, back to the start. The climb is held at $160\pm5$\,bpm; the descent is ridden at self-selected effort. Lap presses at the four gates mark the two timed segments.}
\label{fig:photo_loop}
\end{figure}

\begin{figure*}[t]
\centering
\includegraphics[width=0.9\linewidth]{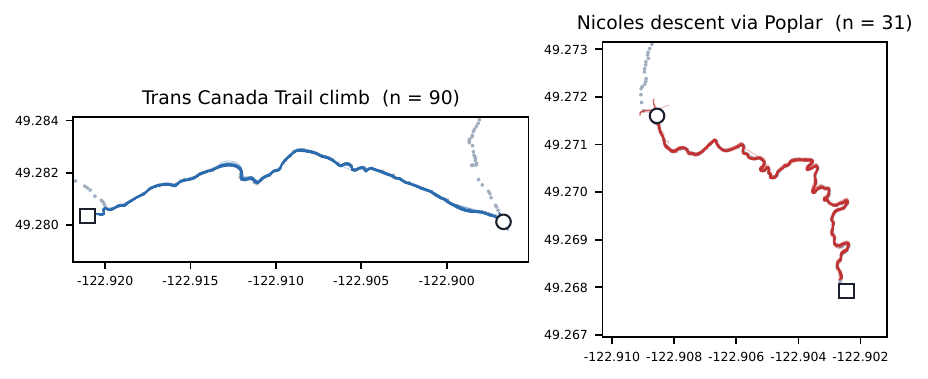}
\caption{\footnotesize Instrument segments on Burnaby Mountain, all traversals to 22~October 2026. \textbf{Left}: $90$ GPS traces on the Trans Canada Trail climb (blue), start gate (circle) and end gate (square) where the gravel meets University Drive. \textbf{Right}: $31$ traces on the Nicoles descent via Poplar Trail (red); the start gate is the athlete's lap-press point, not the OSM junction node, which sat $49$\,m off the ridden line. Grey dots: OSM corridor geometry used for the corridor check.}
\label{fig:instrument_map}
\end{figure*}

\begin{figure*}[t]
\centering
\includegraphics[width=0.9\linewidth]{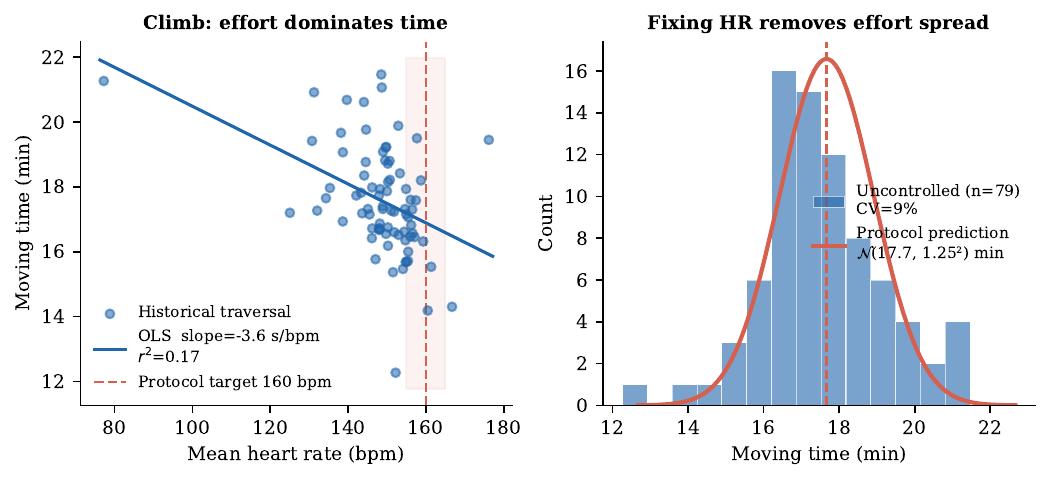}
\caption{\footnotesize Justification for the fixed-heart-rate protocol. \textbf{Left}: moving time vs mean heart rate across $79$ historical TCT traversals; the $-3.6$\,s/bpm slope shows effort explains most between-ride spread. The red band marks the $160\pm5$\,bpm target. \textbf{Right}: uncontrolled time distribution (CV\,9\%) and the protocol prediction $\mathcal{N}(17.7,1.25^2)$\,min; fixing heart rate is expected to collapse the spread to the model's residual uncertainty.}
\label{fig:fixed_hr}
\end{figure*}

\paragraph{Registry protocol.}
The registry is a relational store with seven tools: \texttt{register}, \texttt{predict}, \texttt{record\_evidence}, \texttt{refine}, \texttt{merge}, \texttt{retire}, \texttt{list}. Refinement and merging are permitted only for parents that have reached a verdict or carry at least one piece of evidence, so untested proposals cannot spawn descendants. Every call is logged with its caller, so the audit trail distinguishes an LLM proposal from a tool verdict. Schema and prompts are in Appendix~\ref{app:registry}.

%%%%%%%%%%%%%%%%%%%%%%%%%%%%%%%%%%%%%%%%%%%%%%%%
\section{Evaluation}
\label{sec:eval}

\subsection{Data}
\label{sec:data}
The subject is one recreational cross-country mountain biker (an author) whose Garmin record begins with the purchase of a Forerunner~965 on \DateStart{} and runs to \DateEnd{} (\Span{}). It holds \NActivities{} activities, of which \NRides{} are rides (\NHours{} hours; median ride $79$\,min, $11$\,km, $340$\,m of climbing), recorded with optical wrist heart rate, GPS and a barometric altimeter; no power meter or chest strap was used. Rides are fetched as native FIT files at $1$\,Hz (the vendor's chart endpoint, which downsamples to $2{,}000$ points, is not used); heart-rate coverage within rides is $99.5\%$ and GPS coverage $95.7\%$. Daily wellness records (\NDays{} days) include sleep stages and overnight heart-rate variability ($77\%$ of nights), resting heart rate, stress, body battery and device-computed training metrics (readiness, acute and chronic load, VO$_2$max; $98$--$99\%$ of days). Ambient temperature on rides spans $1$--$32^{\circ}$C with $17$ rides at or above $25^{\circ}$C. Segment discovery yields \NSegments{} segments ($11$ climbs, $8$ descents, $5$ rolling) with \NTraversals{} traversals; \NSegmentsTen{} segments have at least ten traversing rides. Three riding areas dominate: a home hill with $153$ rides, and two mountain networks with $29$ and $26$ rides. The non-ride activities ($265$ strength sessions, hiking, running) contribute to daily load but not to segment outcomes.

\begin{figure*}[t]
\centering
\begin{minipage}[t]{0.49\linewidth}\centering\includegraphics[width=\linewidth]{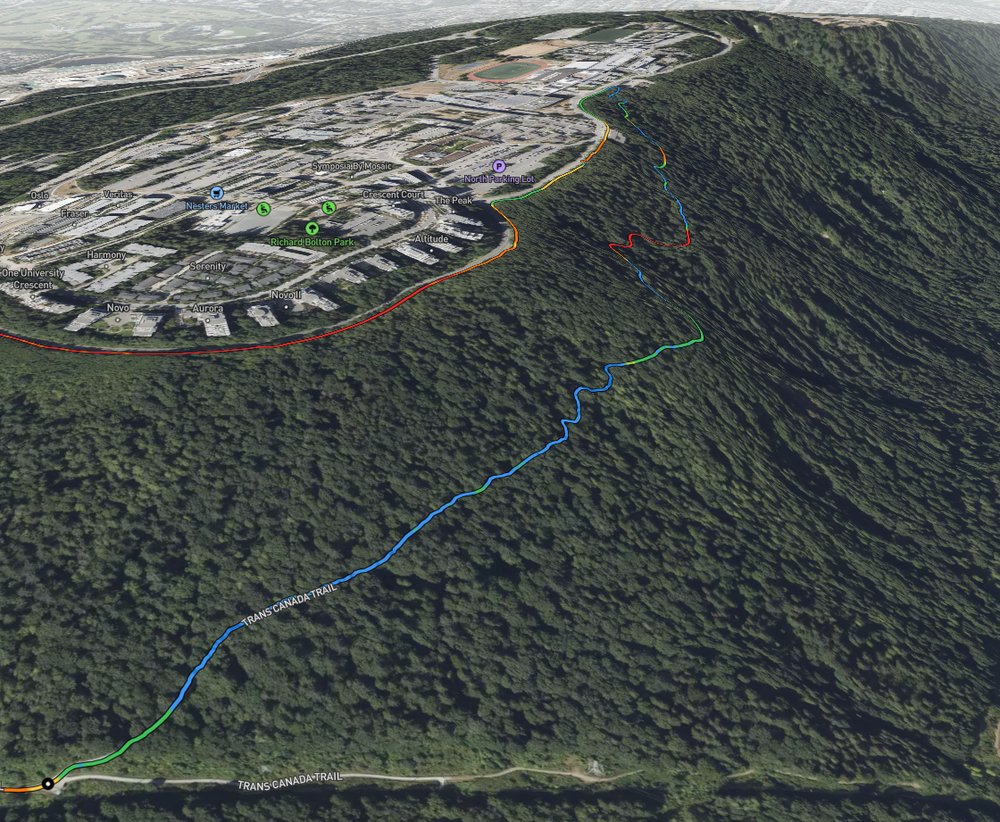}\end{minipage}\hfill
\begin{minipage}[t]{0.49\linewidth}\centering\includegraphics[width=\linewidth]{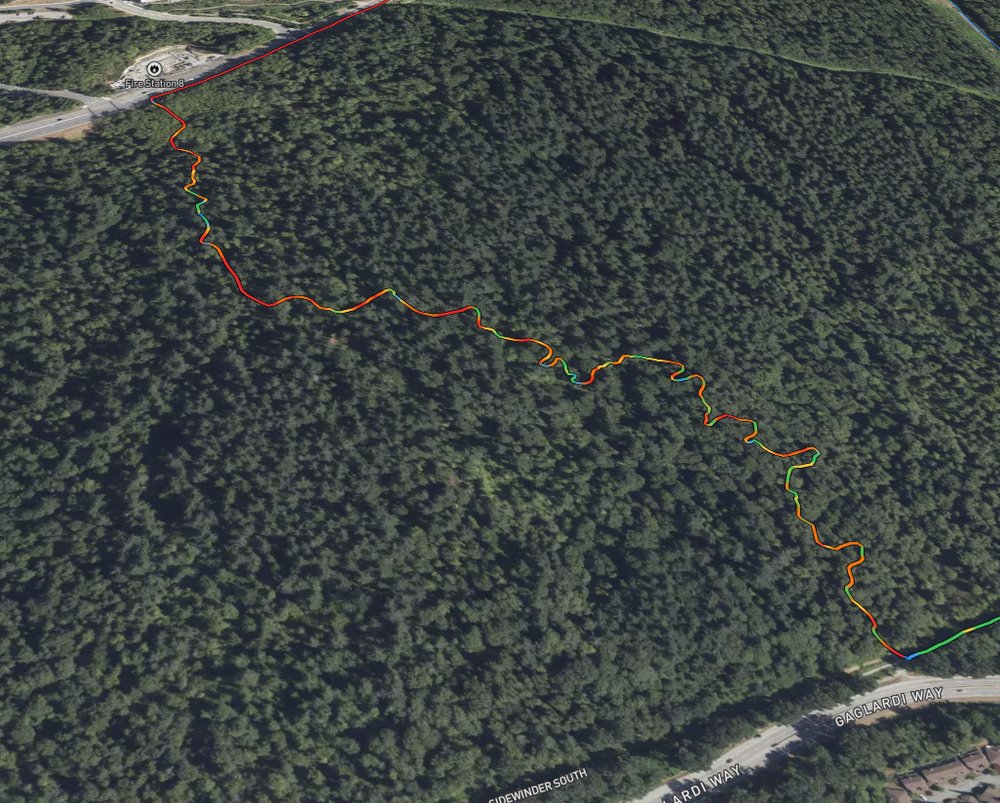}\end{minipage}
\caption{\footnotesize The two instruments in relief (satellite view, one ride, track coloured by heart-rate zone). \textbf{Left}: the Trans Canada Trail climb, $2.06$\,km at $7.2\%$ on wide compacted gravel with no technical features: heart rate is a clean function of mechanical work here, which is why it serves as the fitness instrument. \textbf{Right}: Nicoles Trail, $1.16$\,km and $-158$\,m of tight, rooted singletrack: heart rate decouples from work and time is governed by skill, which is why it is modelled separately (Section~\ref{sec:twin}).}
\label{fig:photo_segments}
\end{figure*}

\paragraph{Instrument ride.}
Automatic discovery keys on named OSM trails and therefore fragments long climbs (the home hill's main gravel ascent is mapped as $22$ OSM ways and does not survive discovery as a single piece). The instrument climb was instead \emph{declared} with the athlete as a gate-to-gate route on the home hill (Fig.~\ref{fig:photo_loop}; close-ups in Fig.~\ref{fig:photo_segments}; GPS traces of every traversal in Fig.~\ref{fig:instrument_map}): the Trans Canada Trail from its lower gate to the point where the gravel meets the summit road, $2.06$\,km, $+148$\,m, $7.2\%$, a wide, smooth, steady-gradient surface with no technical content, $10$ minutes from the athlete's door. Traversals are extracted by gate passage with a corridor check ($\geq 80\%$ of samples within $25$\,m of the trail). Seventy-nine historical traversals (April 2025--September 2026, every month of the record; $90$ including the prospective phase) take $12$--$21$\,min moving (median $17.3$, robust CV $7.4\%$) at mean heart rates of $130$--$176$\,bpm; across those rides each $+1$\,bpm of mean heart rate buys about $3.6$\,s, i.e.\ the between-ride spread is dominated by effort (Fig.~\ref{fig:fixed_hr}), which is precisely what the fixed-heart-rate protocol removes. The descent instrument is Nicoles Trail, entered from the summit via Poplar Trail and timed from the Poplar--Nicoles junction to the exit ($1.16$\,km, $-158$\,m, $-13.6\%$, $23$ traversals, median $5.6$\,min, robust CV $9.2\%$). A longer declared climb on a second network ($5.9$\,km, $+425$\,m, $47$\,min) was considered first and rejected before any prospective ride: it had six traversals against seventy-nine, and an instrument the athlete will ride twice a week has to be near home. The change is logged in the registry (the two ledger predictions registered against the earlier instrument are marked withdrawn, not deleted) and no prospective data is affected.

\subsection{Retrospective replay benchmark}
\label{sec:replay}
The replay asks a single question: at each point in the past \Span{}, would the system's beliefs have \emph{predicted} the following week's rides better than the alternatives? We step a rolling origin $t$ weekly after a four-month warm-up. Every method sees data $\leq t$ only, including language-model calls (frozen model version, temperature $0$, responses cached per origin, three seeds). Each issues predictive distributions for the next seven days of realized rides: (a)~traversal time on each matched segment; (b)~mean climb heart rate conditional on realized virtual power and temperature; (c)~within-ride drift slope; (d)~sign of the weekly change in device VO$_2$max. Registry hypotheses additionally issue binary ledger predictions.

\paragraph{Metrics.}
Continuous ranked probability score and log score for (a)--(c)~\cite{gneiting2007strictly}; probability integral transform histograms and expected calibration error for calibration; Brier score and calibration error for binary ledger predictions; and the posterior entropy of $\theta$ over time as an information-accrual curve.

\paragraph{Baselines.}
(1)~Persistence and exponentially weighted moving averages of segment times; (2)~the device's own VO$_2$max, race-time predictor and training-status outputs, which see identical sensor data; (3)~a Banister model with population constants $(\tau_1,\tau_2)=(42,7)$; (4)~a Gaussian-process regression on calendar time and covariates with no mechanism; (5)~a sequence model for heart-rate forecasting in the style of personalized workout models; (6)~a PHIA-style ReAct agent with a code runtime asked for the same forecasts; (7)~a static, non-agentic twin: our model stack with fixed structure, no registry, no hypothesis-driven refinement.

\paragraph{Ablations.}
Run: no identifiability gate (every Savage--Dickey result admitted); no registry (the twin refit at every origin with fixed structure). The language-model-only forecaster is the ReAct baseline. Not run: no segment instruments (whole-ride outcomes only) and no heart-rate quality control. The active component cannot be ablated in replay, since history is fixed; we say so and evaluate it prospectively.

\paragraph{Statistical comparison.}
Everything is one long dependent series, so we use forecast-comparison tests built for that case: Diebold--Mariano tests on per-origin score differences with heteroskedasticity- and autocorrelation-consistent variance (lag four weeks)~\cite{diebold1995comparing}; moving-block bootstrap with $14$-day blocks for intervals on aggregate differences~\cite{kunsch1989jackknife}; and Giacomini--White tests of conditional predictive ability to characterize \emph{when} the agent helps~\cite{giacomini2006tests}.

\begin{table*}[t]
\centering
\caption{\footnotesize Rolling-origin replay, $60$ weekly origins, one-week horizon. CRPS (lower is better) with $95\%$ moving-block bootstrap intervals; $90\%$-interval coverage; Diebold--Mariano $p$ against the twin row of the same target (HAC variance, lag $4$). Segment time is scored in log-seconds. The ``no segment instruments'' and ``no HR quality control'' ablations listed in Section~\ref{sec:replay} were not run.}
\label{tab:replay}
\footnotesize
\setlength{\tabcolsep}{4pt}
\setlength{\tabcolsep}{3pt}\begin{tabular}{lcccccc}
\toprule
& \multicolumn{3}{c}{Segment time ($n{=}81$)} & \multicolumn{3}{c}{Conditional climb HR [bpm] ($n{=}106$)}\\
\cmidrule(lr){2-4}\cmidrule(lr){5-7}
Method & CRPS & Cov$_{90}$ & DM $p$ & CRPS & Cov$_{90}$ & DM $p$\\
\midrule
Persistence                    & 0.052 [0.047, 0.060] & 0.96 & 0.93 & 9.25 [7.50, 11.06] & 0.88 & $<$0.001\\
EWMA                           & \textbf{0.046} [0.040, 0.055] & 0.96 & 0.31 & 7.87 [6.48, 9.19] & 0.89 & 0.003\\
Ridge on covariates            & 0.109 [0.081, 0.134] & 0.83 & 0.004 & 8.06 [6.69, 10.07] & 0.83 & 0.13\\
GP on time + covariates        & 0.097 [0.070, 0.126] & 0.75 & 0.026 & 8.06 [6.88, 9.46] & 0.70 & 0.004\\
Banister, population $\tau$    & 0.056 [0.049, 0.064] & 0.91 & 0.21 & --- & --- & ---\\
\midrule
Static twin (FFM / HR model) & 0.048 [0.041, 0.063] & 0.91 & --- & \textbf{7.03} [6.13, 7.81] & 0.87 & ---\\
\sys{} agentic twin (gated)   & 0.050 [0.042, 0.065] & 0.91 & 0.007 & 7.00 [6.07, 7.82] & 0.88 & 0.627\\
\quad no identifiability gate  & 0.050 [0.042, 0.065] & 0.91 & 0.007 & 7.00 [6.07, 7.82] & 0.88 & 0.627\\
\quad no registry (static refit) & 0.050 [0.042, 0.065] & 0.91 & 0.005 & 6.89 [5.98, 7.67] & 0.88 & 0.313\\
ReAct agent (PHIA-style)       & 0.144 [0.085, 0.213] & 0.84 & 0.014 & 8.91 [7.58, 10.26] & 0.73 & $<$0.001\\
\bottomrule
\end{tabular}
\end{table*}

\begin{figure*}[t]
\centering
\includegraphics[width=\linewidth]{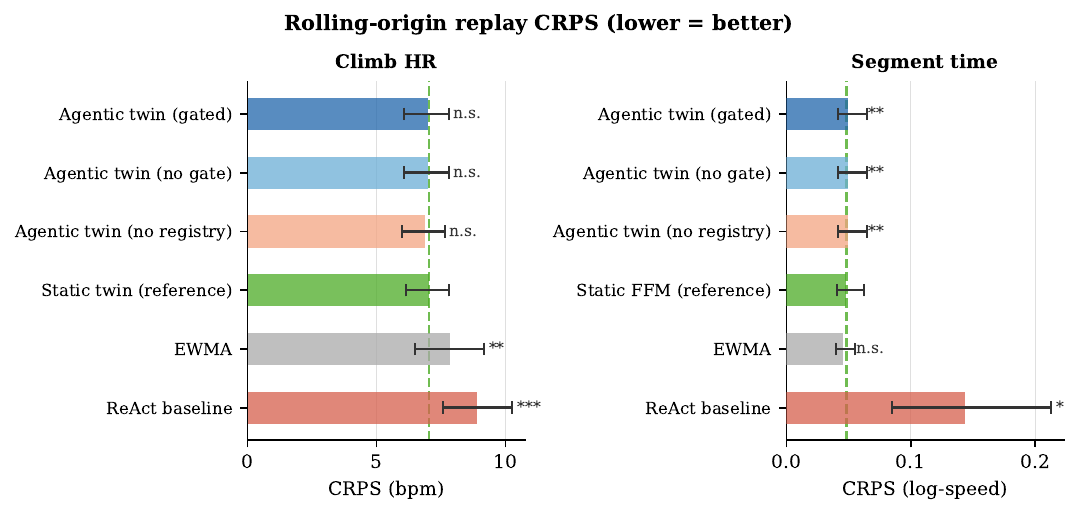}
\caption{\footnotesize Rolling-origin replay CRPS (lower is better) with $95\%$ moving-block bootstrap CIs; DM test significance vs the static twin reference (green dashed line). The agentic system matches the static twin on both targets (n.s.\ on HR, $p{=}0.007$ on segment time); the ReAct baseline is significantly worse on both.}
\label{fig:replay_crps}
\end{figure*}

\subsection{Prospective phase}
\label{sec:prospective}
Since 3~October 2026 the weekly cycle has run live on the athlete. Each week contains two \emph{instrument rides} that never change: the Trans Canada Trail climb held at $160\pm5$\,bpm and the Nicoles descent at self-selected effort, each bracketed by lap presses at the gates, on day~1 (\emph{fresh}, after a rest day) and day~5 (\emph{loaded}, after the week's training). What the agent designs is the training \emph{between} them: one of fifteen one-, two- or three-week load patterns (conventional, front/back-loaded, deload $50/25/0\%$, overload $115/130\%$ and their sequences), chosen by Eq.~\eqref{eq:utility} and checked by the safety officer. Before every instrument ride the ledger holds a sealed prediction for it; a ride is scored by the first traversal after sealing, one traversal scores at most one prediction, and a ride through the gates without the lap protocol is training data and never scores. The twin is refit once per week, after all of that week's predictions have been scored and before the next are sealed, so no prediction is ever made by a model that has seen the ride it predicts. Two further pre-registered components begin once the baseline is established: randomization of weeks to information-directed versus conventional ($\lambda_t{=}0$) design, with realized information per ride as the outcome; and short-washout within-person interventions (sleep extension, time of day, heat, caffeine) in counterbalanced one-week blocks. Design details and power are in Appendix~\ref{app:nof1}.

\subsection{Synthetic recovery check}
\label{sec:synthetic}
A single subject cannot establish generality, and a posterior should not be interpreted before the model has been shown to recover known parameters. We therefore simulate athletes from the fitted twin under the athlete's \emph{actual} training history and refit; with the real design the fitness--fatigue model recovers the daily performance state ($r{=}0.996$) and the load scale $k_1$ and ratio $k_2/k_1$ within their $90\%$ intervals, but not the time constants, whose posteriors remain close to their priors (Appendix~\ref{app:synthetic}). A multi-athlete simulator for information-accrual experiments is future work.

\begin{figure*}[t]
\centering
\includegraphics[width=\linewidth]{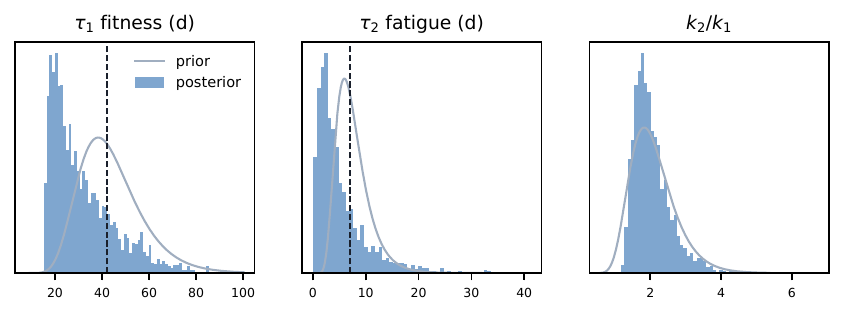}
\caption{\footnotesize Posterior distributions from the retrospective twin. \textbf{Left}: cardiac drift $\gamma_d$ is clearly positive (H1, belief $95\%$). \textbf{Centre}: fitness time constant $\tau_1$ remains wide despite $79$ instrument traversals (H4, belief $80\%$); the prospective fresh/loaded pair is designed to resolve this. \textbf{Right}: switching from $6$ Mountain Highway rides to $79$ TCT rides tightened HR model noise from $6.5$ to $5.3$\,bpm.}
\label{fig:posteriors}
\end{figure*}

\subsection{Results}
\label{sec:results}
Table~\ref{tab:replay} and Figure~\ref{fig:replay_crps} report the replay over $60$ weekly origins after a $120$-day warm-up; a segment becomes a forecast target once it has three prior traversals. The corresponding twin posteriors are in Figure~\ref{fig:posteriors}.

\paragraph{Conditional heart rate.}
The twin's heart-rate response model attains CRPS $7.03$\,bpm ($95\%$ block-bootstrap interval $6.13$--$7.81$) against $7.87$ for an exponentially weighted average, $8.06$ for the Gaussian process, $8.06$ for ridge regression on the same covariates and $9.25$ for persistence; Diebold--Mariano tests on per-origin losses give $p=0.003$, $0.004$, $0.13$ and $<0.001$ respectively. Its $90\%$ intervals cover $86.8\%$ of outcomes (ECE $0.15$); the model is slightly overconfident, which we attribute to ride-level variation that a one-week-ahead forecast cannot condition on.

\begin{figure*}[t]
\centering
\includegraphics[width=0.78\linewidth]{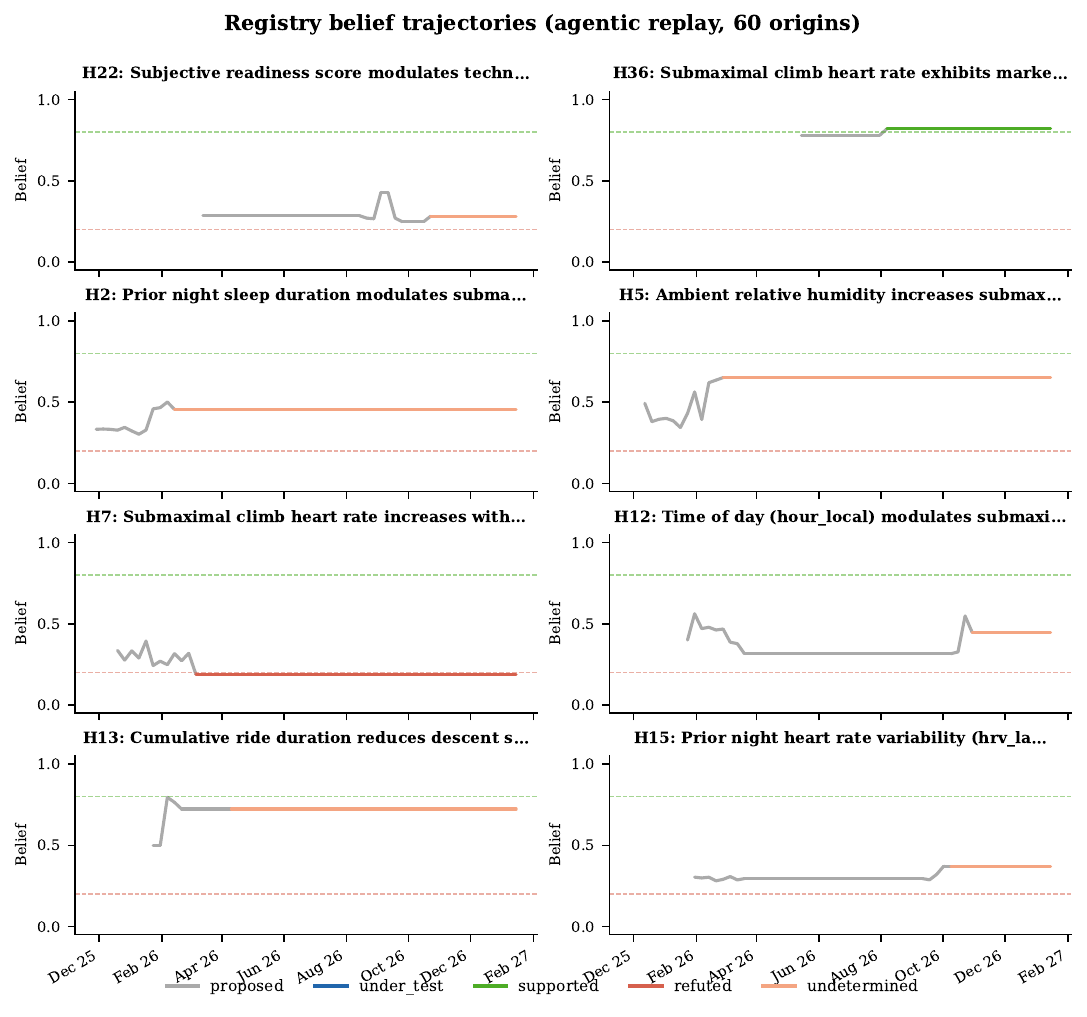}
\caption{\footnotesize Belief trajectories for the eight most-tested hypotheses in the agentic replay ($60$ rolling origins). Colour encodes state: grey\,=\,proposed, blue\,=\,under test, green\,=\,supported, red/orange\,=\,refuted/undetermined. Flat segments show hypotheses that reached a verdict; invariant I8 (no permanent verdicts) would reopen them if later evidence disagreed.}
\label{fig:registry}
\end{figure*}

\paragraph{Agentic replay.}
With the fixed registry (invariants I7--I8), the hypothesis loop runs causally at every origin: base twin refit, Hypothesizer + Modeler/Critic against a frozen-clock registry, Statistician evaluates open hypotheses (directional claims by Savage--Dickey on the stored fit; structural specs by compiling a linear term and refitting), working model = base + admitted structural specs (belief $\geq 0.5$). Three ablations share the same base fits and differ only in admission: \emph{gated} (identifiability gate on), \emph{no-gate}, \emph{no-registry} (static twin). Across 60 origins the agentic system matches the static twin on HR forecasting (CRPS $7.00$ vs $7.03$, $p=0.63$) and segment time ($0.050$ vs $0.048$, $p=0.007$, marginally worse). The no-registry ablation is indistinguishable. I7 reduced the hypothesis count from 62 to 40 by blocking duplicate registrations; I8 demoted 5 premature ``supported'' verdicts when later data disagreed (belief trajectories in Figure~\ref{fig:registry}). The honest interpretation: 18 months of observational data, with the current segment network, does not give the hypothesis loop enough contrast to improve forecasts. The designed experiments in the prospective phase (particularly the fresh/loaded pair and the overload--rest contrast) are what the loop needs.

\paragraph{ReAct baseline.}
A PHIA-style agent (the same language model with a read-only SQL tool over the same history, asked to reason and return a Gaussian forecast) scores $8.91$\,bpm on conditional heart rate and $0.144$ on segment time, significantly worse than the twin on both ($p<0.001$, $p=0.014$) and worse than persistence on segment time. Free-form reasoning over the raw table does not recover what the mechanistic model and even a moving average extract; this is the gap the architecture is designed around.

\begin{figure*}[t]
\centering
\includegraphics[width=\linewidth]{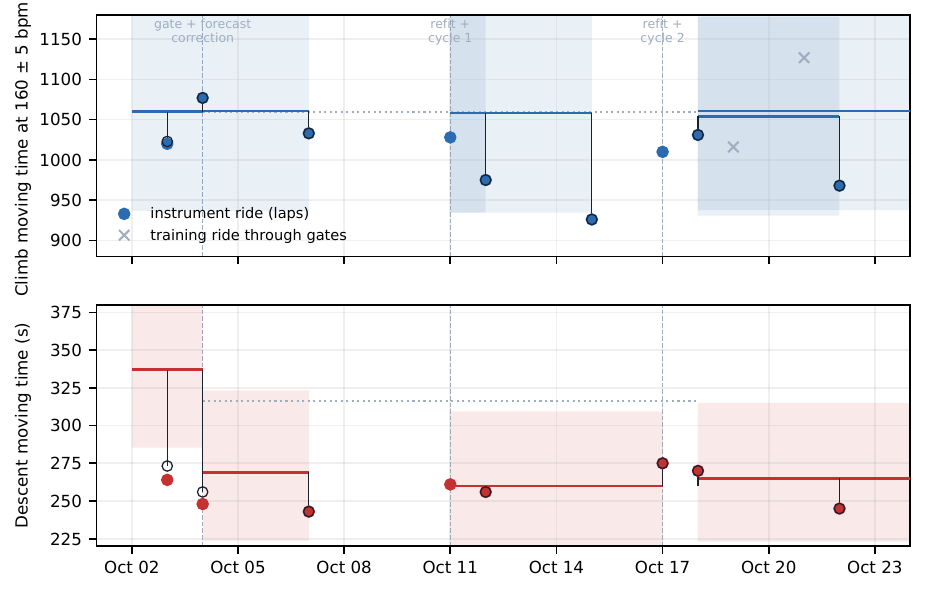}
\caption{\footnotesize The prospective ledger, 1--23~October 2026. Horizontal bars are sealed predictions (mean; shaded $90\%$ interval) drawn from the moment of sealing to the ride that scored them; dotted bars are predictions withdrawn before any outcome (I9). Filled circles are instrument rides (lap protocol), open circles the outcome as scored, crosses training rides through the gates (not scored). Dashed verticals: the descent-gate and forecast correction (4~Oct) and the two live refit-and-cycle events. \textbf{Top}: the climb forecast barely moves while the athlete's time at $160$\,bpm falls by $\sim\!80$\,s. \textbf{Bottom}: descent forecasts before and after the instrument correction.}
\label{fig:prospective}
\end{figure*}

\paragraph{Segment time.}
No method beats a simple exponentially weighted average of past traversal times (CRPS $0.046$ in log-seconds, i.e.\ ${\sim}4.6\%$): the fitness--fatigue twin with learned time constants scores $0.048$, persistence $0.052$ and a Banister model with population constants $0.056$, none significantly different from one another, while the Gaussian process and ridge baselines are clearly worse ($0.097$, $0.109$). This is the forecasting face of the identifiability result of Section~\ref{sec:ffm}: in this athlete's observational record, training load carries no predictive information about segment time beyond recent performance. We regard this as a finding, not a failure, and as the strongest available motivation for the designed contrasts of the prospective phase.

\paragraph{Prospective phase: three weeks.}
Figure~\ref{fig:prospective} shows every sealed prediction and every instrument ride from 3 to 22~October 2026: $20$ predictions sealed, $14$ scored, $4$ withdrawn before their outcome existed (I9), $2$ open. The athlete rode $9$ protocol climbs at $160$--$164$\,bpm (one at $170$) and $8$ protocol descents (one while no descent prediction was open); two further climbs through the gates without laps were training and were not scored.

\emph{Descent.} The first two descent predictions were sealed against a gate that sat on the OSM junction node, $49$\,m from the line the athlete rides; both outcomes fell $1.9$ and $2.4\sigma$ below the forecast (CRPS $0.155$, $0.218$ log-s). They were scored as sealed, the gate was moved to the athlete's lap-press point, the correction was logged as a registry event, and the forecast was changed from a historical median to the skill-model posterior. The five predictions since score CRPS $0.040$ log-s ($4.0\%$) with all standardized residuals inside $\pm0.9$: the descent instrument is calibrated.

\emph{Climb.} Seven climb predictions scored a mean CRPS of $40$\,s ($3.9\%$ of the $17$-minute climb) with $90\%$-interval coverage $6/7$, but the athlete was faster than the forecast in six of seven, mean standardized residual $-0.71$. Across the three weeks the protocol climb time fell from $1046$\,s (first three) to $1003$\,s (last three) at matched heart rate, a slope of $-4.8$\,s/day; the twin's forecast moved from $1060$ to $1054$\,s. An exponentially weighted average of past climbs at $155$--$167$\,bpm, sealed at the same moments, would have scored $31.5$\,s. This is the live counterpart of the replay's segment-time result: the fitted fitness--fatigue state is too slow to track a real training response, and the pre-specified standard for the twin (match or beat the moving average) was not met.

\begin{table*}[t]
\centering
\caption{\footnotesize Registry state on 22~October 2026 after three live cycles. Beliefs start at $0.5$ and move only through Savage--Dickey evidence from stored fits, subject to the identifiability gate. H1--H5 were pre-specified; H6--H11 were registered by the agent during the live cycles. Ledger: predictions scored / sealed per hypothesis, with mean CRPS (climb in seconds, descent in log-seconds).}
\label{tab:registry}
\vspace{2pt}
\footnotesize
\resizebox{\linewidth}{!}{%
\begin{tabular}{llccl}
\toprule
Hypothesis & Formal object & Belief & State & Ledger\\
\midrule
H1 Cardiac drift $>0$ & $\gamma_d>0$ & 0.95 & under test & 3/3 scored, CRPS 21.9\,s\\
H2 Drift larger in heat & $\gamma_{dT}>0$ & 0.21 & proposed & ---\\
H3 Less sleep slows descents & $\delta_{\mathrm{sleep}}<0$ & 0.94 & under test & 7/7 scored, CRPS 0.082 (0.040 post-correction)\\
H4 Fitness constant $<42$\,d & $\tau_1<42$ & 0.50 & under test (gated) & 4/4 scored, CRPS 54.1\,s; falsifier fired $2/2$\\
H5 Aerobic response improved & $c_{w_{\mathrm{last}}}-c_{w_{\mathrm{first}}}<0$ & 0.14 & proposed & ---\\
H6 Climb HR rises with temperature & $\gamma_T>0$ & 0.43 & proposed & ---\\
H7 Wet trail slows descents & $\delta_{\mathrm{wet}}>0$ & 0.65 & proposed & ---\\
H8 Climb HR varies with time of day & $\delta_{\mathrm{hour}}\neq0$ & 0.62 & proposed & ---\\
H9 Descent slows with elapsed ride time & $\delta_{\mathrm{elapsed}}>0$ & 0.49 & proposed & ---\\
H10 Fatigue constant $<7$\,d & $\tau_2<7$ & 0.55 & proposed & ---\\
H11 Sleep affects climbing HR & $\delta_{\mathrm{sleep}}^{\HR}\neq0$ & 0.36 & proposed & ---\\
\bottomrule
\end{tabular}}
\end{table*}

\emph{Pre-registered contrasts.} Both live cycles sealed a fresh and a loaded climb prediction and stated the falsifier in advance: a loaded test not slower than the fresh one counts against the modelled fatigue penalty. In week~1 the fresh test took $975$\,s and the loaded test $926$\,s; in week~2, $1031$\,s ($6.0$\,h sleep) and $968$\,s. The falsifier fired twice. The refit that followed moved the fitness--fatigue posterior to $\tau_1=27$\,d $[17,59]$, $P(\tau_1<42)=0.80$, and $\tau_2=3.7$\,d $[0.7,15.9]$; the identifiability gate still refuses to move H4's belief from $0.5$ because the prior--posterior overlap is $0.84$. Two weeks of designed contrast have not resolved what eighteen months of observation could not, and the registry says so.

\emph{The agent and the safety officer.} In both live cycles the design ranked first by expected information gain was an overload week ($130\%$ of the four-week mean); the safety officer vetoed it both times (ACWR $1.35>1.3$; weekly load outside $\pm15\%$) and the agent took the second-ranked two-week plan. The $\pm15\%$ tolerance made the pre-registered overload design logically infeasible. With the athlete's written consent the cap was raised to ACWR $1.5$ and $\pm35\%$ and a no-consecutive-overload rule added, effective from the fourth cycle; the change is a \texttt{protocol\_change} event in the registry and is not retroactive. The agent also registered six new hypotheses (heat, time of day, wet surface, elapsed time, $\tau_2<7$\,d, sleep on climbing heart rate); all remain \emph{proposed} with $|\log\mathrm{BF}|<0.8$.

\emph{Audit trail.} The registry holds $4$ \texttt{score\_correction} events from this period: three cases in which an ad-hoc scorer matched the wrong ride (date-level instead of timestamp comparison; one traversal claiming two predictions; a training ride claiming a test prediction), each reversed with the wrong value preserved, and the rules that now prevent them (first traversal strictly after sealing; invariant I10, one observation scores one prediction; lap protocol required). We report these because a system whose record is clean on first contact with reality is a system whose record is not being kept. Full registry state, including the six agent-proposed hypotheses, is in Table~\ref{tab:registry}.

%%%%%%%%%%%%%%%%%%%%%%%%%%%%%%%%%%%%%%%%%%%%%%%%
\section{Limitations and Ethics}
\label{sec:limitations}

\paragraph{One subject.}
The empirical study is $N{=}1$ by design, and the subject is an author. We argue generality on methodological grounds (the formalization, the benchmark, the simulator and a pipeline that admits a second athlete without code changes), not on the basis of the single subject's physiology. Replay evaluates only the passive component of the system; the active component is evaluated prospectively over a short horizon with limited power for small effects, and we state in advance which effects are detectable.

\paragraph{Sensors.}
Wrist optical heart rate and forest GPS are noisy; the quality-control and Student-$t$ likelihoods mitigate but do not remove this. Virtual power is a modelling input, not a measurement. Several device metrics used as baselines are proprietary; the retrospective twin does not use them as inputs, and if the auxiliary VO$_2$max channel is enabled we will report results with and without it.

\paragraph{Identifiability.}
The fitness--fatigue component is poorly identified after \Span{} of observational data (Section~\ref{sec:ffm}) and may remain so after a short prospective block. The system is designed to say so.

\paragraph{Self-experimentation and safety.}
The prospective phase is self-experimentation by a healthy adult with hard caps on heart rate and load, readiness and illness gates, and interventions restricted to sleep, timing, heat exposure and caffeine within ordinary ranges. \sys{} is a research prototype and not medical advice; nothing in the design should be read as safe for individuals with cardiovascular or other conditions without clinical oversight. Garmin's Connect API is unofficial; raw responses are snapshotted so that the study is reproducible against API drift.

\paragraph{Three weeks.}
The prospective results are three weeks of one athlete: $14$ scored predictions and two fresh/loaded contrasts. They are enough to show the ledger working, including against the twin, and not enough to estimate a time constant. The randomized information-directed versus conventional comparison has not yet begun; every live week so far was information-directed. We also changed the instrument (gate position), the descent forecast method and the safety thresholds during the phase. Each change is dated and logged, predictions sealed before a change were scored under the rules in force when sealed, and none was retroactive; but a reader should weigh the first three weeks as a shakedown of the protocol as much as a test of the athlete.

\paragraph{Language-model nondeterminism.}
Model versions are frozen, temperature is zero, every call is cached per origin, and spread over three seeds is reported.

%%%%%%%%%%%%%%%%%%%%%%%%%%%%%%%%%%%%%%%%%%%%%%%%
\section{Conclusion}
\label{sec:conclusion}
Personal-health agents have learned to read wearable data and to speak about it well. We have argued that the next step is not better speech but accountable learning: a system that holds explicit, probabilistic claims about one person, commits to predictions before seeing outcomes, chooses what to observe next, and is judged by whether it was right. \sys{} realizes this loop for a mountain biker with nothing but a watch, with a protocol that lets language models do what they are good at (propose mechanisms, choose instruments, negotiate with a human, explain) while deterministic tools fit, score, gate and record. Eighteen months of wrist data did not resolve a fitness time constant, and three weeks of designed rides have so far shown the twin to be too slow to follow the athlete it models; a system that writes that down, keeps the wrong scores next to the corrected ones, and can answer ``not yet, and here is the week that would help'' is, we think, the right kind of coach.

%%%%%%%%%%%%%%%%%%%%%%%%%%%%%%%%%%%%%%%%%%%%%%%%
\section*{Code and Data Availability}
The \sys{} implementation, the full hypothesis registry and prediction ledger, the replay benchmark, and the operational protocol are released at \url{https://github.com/Bmoradi93/epicoach-agent}.
\bibliographystyle{IEEEtran}
\bibliography{references}

%%%%%%%%%%%%%%%%%%%%%%%%%%%%%%%%%%%%%%%%%%%%%%%%
\appendices

\section{Segment Discovery}
\label{app:segments}

\paragraph{Trail network.}
For each riding area (a bounding box around a cluster of ride start points) we query OpenStreetMap for ways with a \texttt{highway} tag in \{path, track, cycleway, footway, bridleway, service, unclassified, steps\} and a \texttt{name}; responses are cached. Ways sharing a name are greedily joined when endpoints lie within $15$\,m, then densified to $5$\,m. Chains shorter than $100$\,m are dropped. Paved roads (\texttt{unclassified}) are retained for segment discovery but excluded from the heart-rate and fitness--fatigue models, where virtual power is unreliable at road speeds.

\paragraph{Snapping and runs.}
A KD-tree over all chain points returns, for each GPS sample, the nearest chain and its along-trail coordinate $s$, accepted if within $20$\,m. Gaps of up to $30$\,s on a chain are bridged. A run needs $\geq 30$ samples, $\geq 20$ snapped, and a robust slope $|\dd s/\dd t|\geq 0.3$\,m/s (otherwise the rider is dithering at a junction).

\paragraph{Consensus gates.}
Chain length is binned at $10$\,m; each ride's coverage interval $[\min s,\max s]$ increments the bins it spans; the segment is the longest contiguous bin range with count $\geq\max(5,\,0.6\times\text{rides touching the chain})$ and length $\geq 250$\,m. Segments are typed by the median altitude change over traversals (climb $>15$\,m, descent $<-15$\,m, otherwise rolling).

\paragraph{Traversals.}
Within a run, entry is the first crossing of the entry gate (a run that starts more than $30$\,m inside the segment is not a full traversal), exit the first subsequent crossing of the exit gate. Accepted if $\geq 85\%$ of samples lie within the gates $\pm15$\,m, no inter-sample gap exceeds $90$\,s, progress is monotone for $\geq 90\%$ of steps, and duration $\geq 60$\,s. Laps yield multiple traversals per run.

\paragraph{Known failure modes.}
GPS error under canopy of $10$--$25$\,m can place a sample on a parallel trail; the $20$\,m snap radius and the monotone-progress test suppress most such errors. Trails absent from OpenStreetMap, or unnamed, produce no segments; in this study $20\%$ of ride samples in the main area were unmatched, mostly unsanctioned connectors. Entry/exit jitter of ${\sim}1$\,s is $2\%$ of a $60$\,s descent, so we prefer segments of several minutes for hypothesis testing and report duration alongside every estimate.

\paragraph{Per-traversal record.}
{\raggedright\texttt{segment\_id, activity\_id, t\_entry, t\_exit, duration\_s, moving\_s, has\_stop, cover, hr\_mean, hr\_max, hr\_ok\_frac, d\_alt, dist\_m, elapsed\_at\_entry\_s, vp\_mean\_w, grade\_mean, wx\_temp\_c, wx\_humidity, wet\_now, sleep\_h, hrv\_last\_night, readiness\_score, week, day\_index}.\par}

\section{Model Specifications and Priors}
\label{app:models}

\paragraph{Heart-rate response (Eqs.~\ref{eq:hrlag}--\ref{eq:hrss}).}
$\tau_{hr}\sim\mathrm{LogNormal}(\ln 30,0.3)$\,s; $\alpha_r\sim\mathcal{N}(\mu_\alpha,\sigma_\alpha^2)$ with $\mu_\alpha\sim\mathcal{N}(120,15^2)$, $\sigma_\alpha\sim\mathrm{HalfNormal}(8)$; $\beta_{w}\sim\mathcal{N}(\beta_{w-1},\sigma_\beta^2)$, $\beta_0\sim\mathcal{N}(0.25,0.1^2)$ bpm/W, $\sigma_\beta\sim\mathrm{HalfNormal}(0.01)$; $\gamma_T\sim\mathcal{N}(0.5,0.5^2)$ bpm/$^{\circ}$C; $\gamma_d\sim\mathcal{N}(4,3^2)$ bpm/h; $\gamma_{dT}\sim\mathcal{N}(0,0.5^2)$; $u_s\sim\mathcal{N}(0,\sigma_u^2)$, $\sigma_u\sim\mathrm{HalfNormal}(6)$; $\nu\sim\mathrm{Gamma}(2,0.1)$ truncated below at $2$; $\sigma_{hr}\sim\mathrm{HalfNormal}(3)$. The recursion is implemented with \texttt{jax.lax.scan} on $5$\,s-downsampled series. The segment-aggregate variant replaces Eq.~\eqref{eq:hrlag} by $\overline{\HR}_{r,s}=\HR^{ss}_{r,s}+\epsilon_{r,s}$ with the same linear predictor evaluated at traversal means.

\paragraph{Skill state (Eq.~\ref{eq:skill}).}
$\mu_s\sim\mathcal{N}(\log\text{median}_s,0.2^2)$; $\lambda_s\sim\mathrm{HalfNormal}(0.1)$; $S_0=0$, $\sigma_S\sim\mathrm{HalfNormal}(0.02)$; $\delta_\cdot\sim\mathcal{N}(0,0.05^2)$; $\eta\sim t_\nu(0,\sigma_\eta)$, $\sigma_\eta\sim\mathrm{HalfNormal}(0.05)$.

\paragraph{Fitness--fatigue (Eq.~\ref{eq:ffm}).}
Load: Banister exponential TRIMP $w=\sum_t\Delta t\cdot\HR^{r}_t\cdot 0.64\,e^{1.92\,\HR^{r}_t}$ with heart-rate reserve fraction $\HR^{r}_t=(\HR_t-\HR_{\mathrm{rest}})/(\HR_{\max}-\HR_{\mathrm{rest}})$; $\HR_{\max}$ from the $99.5$th percentile over all rides, $\HR_{\mathrm{rest}}$ from the device's nightly value. Priors: $\tau_1\sim\mathrm{LogNormal}(\ln 42,0.3)$, $\tau_2\sim\mathrm{LogNormal}(\ln 7,0.4)$, $k_1\sim\mathrm{HalfNormal}(0.05)$, $k_2/k_1\sim\mathrm{LogNormal}(\ln 2,0.3)$, $p_0\sim\mathcal{N}(0,1)$ on standardized performance. Measurement: $\log v_{r,s}\mid\HR\text{-matched}=a_s+b\,p_{d(r)}+\boldsymbol{\delta}^{\top}\mathbf{c}_r+\epsilon$, and $\mathrm{VO_2max}^{\mathrm{dev}}_d=c_0+c_1p_d+\epsilon'$ with $\epsilon'$ allowed to be heavy-tailed because device VO$_2$max updates are discrete. Model comparison between $\tau$ hypotheses uses bridge-sampling Bayes factors and PSIS-LOO~\cite{vehtari2017practical}; we report prior--posterior overlap (Bhattacharyya coefficient) for every parameter.

\paragraph{Weekly updates.}
The twin is refit by NUTS once per week, after all predictions for the week are scored and before new ones are sealed; between refits the stored posterior is used as is. Ledger predictions are drawn from it: the climb from the fitness--fatigue posterior at $160$\,bpm (segment level, last daily state, heart-rate slope, season), the descent from the skill posterior with the skill state advanced one random-walk step and covariates at protocol conditions.

\paragraph{Nested Monte Carlo.}
With $\theta^{(n)}\sim p_t(\theta)$, $n=1..N$, and $y^{(n)}\sim p(y\mid\theta^{(n)},d)$,
\begin{equation*}
\begin{split}
\widehat{\EIG}_\theta(d)=\frac1N\sum_{n=1}^{N}\Big[&\log p(y^{(n)}\mid\theta^{(n)},d)\\
&-\log\frac1M\sum_{m=1}^{M}p(y^{(n)}\mid\theta^{(m)},d)\Big],
\end{split}
\end{equation*}
with $N=M=500$ and the inner samples shared across designs~\cite{rainforth2018nesting}. The estimator is biased upward at finite $M$; because we only rank designs within a menu, the bias is tolerable, and we report the rank stability under $M\in\{250,500,1000\}$. For the hypothesis-level quantity, $y$ is scored under each sub-model $M_H$ and $\pi\mid y$ follows from Bayes' rule over the registry.

\paragraph{Menu $\calD_t$.}
Archetypes: (i)~long low-intensity ride, $90$--$180$\,min, heart rate below a zone-2 cap; (ii)~climb repeats on segment $s$ at heart-rate target $h\in\{\text{Z3},\text{Z4}\}$, $3$--$5$ repeats; (iii)~descent laps on segment $s'$, $4$--$8$ laps, easy climbs between; (iv)~deload: $\leq50\%$ of mean weekly load; (v)~standardized test: eight-minute climb on a designated segment at a fixed heart-rate target preceded by a fixed warm-up; (vi)~rest. Each archetype is crossed with day-of-week and time-of-day options drawn from the athlete's stated availability and the weather forecast; the Experiment Designer prunes to $20$--$40$ candidates before scoring.

\paragraph{Feasibility set.}
$\ACWR_t=\overline{w}_{7}/\overline{w}_{28}\in[0.8,1.3]$ after the candidate is appended; $\sum_{\text{week}}w\in[0.85,1.15]\times$ block target; $\geq1$ rest day in any $7$-day window; heart-rate target $\leq0.92\,\HR_{\max}$; device readiness $\geq30$; no athlete-reported illness or injury in the daily check-in; candidate day marked available.

\begin{table*}[t]
\centering
\caption{\footnotesize Discovered segments. Area A is the home hill (instrument network), B and C two mountain networks. Declared routes are gate-to-gate instruments; the rest are discovered. Moving time is the median over traversals; CV is the coefficient of variation of moving time.}
\label{tab:segments}
\scriptsize
\setlength{\tabcolsep}{4pt}
\begin{tabular}{llrrrrrrr}
\toprule
Trail (OSM name) & Area & Kind & Length [m] & $\Delta$alt [m] & Rides & Moving [min] & CV [\%] & HR\\
\midrule
\textbf{TCT, lower gate $\to$ summit road (declared)} & A & climb & 2061 & +148 & 90 & 17.2 & 9.1 & 150\\
\textbf{Nicoles via Poplar (declared)} & A & descent & 1159 & -158 & 31 & 4.8 & 14.8 & 123\\
Dead Moped Trail & A & climb & 692 & +28 & 86 & 4.6 & 11.5 & 140\\
Upper Cut Trail & A & descent & 353 & -51 & 38 & 1.2 & 25.8 & 120\\
Arts Road & A & rolling & 339 & +12 & 31 & 1.6 & 18.5 & 141\\
North Road Trail & A & rolling & 430 & -13 & 30 & 1.9 & 18.3 & 127\\
Sidewinder South & A & climb & 1074 & +96 & 29 & 10.9 & 8.6 & 147\\
North Road & A & climb & 365 & +19 & 27 & 2.4 & 35.3 & 140\\
Sidewinder & A & climb & 971 & +81 & 27 & 9.9 & 6.2 & 146\\
Sidewinder & A & descent & 971 & -76 & 24 & 3.0 & 11.9 & 133\\
Sidewinder South & A & descent & 1070 & -92 & 18 & 2.7 & 15.6 & 117\\
Mountain Air Trail & A & rolling & 1349 & +9 & 18 & 7.0 & 9.4 & 156\\
Cougar Creek Trail & A & rolling & 430 & +11 & 18 & 2.1 & 12.5 & 149\\
North Road Trail & A & descent & 487 & -49 & 16 & 2.5 & 8.7 & 124\\
North Road & A & climb & 755 & +20 & 14 & 4.4 & 12.9 & 124\\
Pipeline Trail & A & climb & 650 & +26 & 11 & 4.8 & 8.8 & 146\\
Lower Snake Trail & A & descent & 519 & -72 & 7 & 2.2 & 8.0 & 128\\
North Road Trail & A & climb & 487 & +47 & 5 & 5.1 & 3.7 & 144\\
Grouse Mountain Highway & B & climb & 840 & +61 & 10 & 7.7 & 6.3 & 152\\
Access road, lot $\to$ 7th switchback (declared, superseded) & B & climb & 5896 & +425 & 6 & 47.3 & 8.9 & 157\\
Skid Road Ascent & B & climb & 469 & +49 & 9 & 4.8 & 6.5 & 152\\
Ramble On & B & rolling & 402 & +6 & 8 & 3.4 & 9.8 & 150\\
Seventh Secret & B & descent & 1028 & -160 & 7 & 7.5 & 6.2 & 103\\
Executioner Ascent & B & climb & 362 & +46 & 7 & 5.0 & 3.3 & 160\\
For the Kids & B & climb & 519 & +60 & 7 & 6.1 & 10.3 & 141\\
Expresso & B & descent & 1490 & -207 & 5 & 6.9 & 3.7 & 100\\
Empress Bypass & C & descent & 510 & -62 & 6 & 1.7 & 2.4 & 128\\
\bottomrule
\end{tabular}
\end{table*}

\section{Prospective N-of-1 Design and Power}
\label{app:nof1}

\paragraph{Randomization.}
Weeks are the unit for the information-accrual comparison, randomized in pairs so that each fortnight contains one information-directed and one conventional week. Within each week, two instrument rides are scheduled on non-adjacent days; the intervention order (AB or BA) is randomized with counterbalancing across weeks. Sleep extension requires a $1$-day washout; caffeine a $48$-hour abstinence baseline; time-of-day/heat manipulations are constrained by forecast.

\paragraph{Instrument ride.}
Two instrument rides per week on the declared climb, at a fixed heart-rate target of $160\pm5$\,bpm chosen by the athlete as sustainable for the $\sim\!17$\,min effort (watch alert on the band; lap press where the gravel meets the summit road ends the step), followed by one traversal of the designated descent at self-selected effort. The two rides occupy fixed positions: a \emph{fresh} test on the first day after the weekly rest day and a \emph{loaded} test on day~5, after the week's training. The pair is the within-week fatigue contrast the fitness--fatigue model needs to separate $\tau_2$ from $\tau_1$ without special weeks; with one ride a week that contrast is unavailable. Everything between the tests is the athlete's normal training (rides, football, strength work; all enter the load series, with the caveat that heart-rate-based load under-counts strength sessions). Outcomes: moving time on the climb at the protocol heart rate, descent moving time. Each instrument ride carries $\approx 60$ exp-TRIMP, about $12\%$ of the athlete's weekly load, so the plans in the design menu differ only in the remaining training.

\paragraph{Estimand and analysis.}
$\psi=\E[\log v^{(1)}-\log v^{(0)}]$ within person. Primary analysis: Bayesian regression of $\log v$ on treatment, block effects, a linear trend and AR(1) residuals; secondary: randomization test permuting treatment labels within blocks; sensitivity: time-varying g-formula with one-step carryover~\cite{daza2024causal}.

\paragraph{Power.}
With traversal-speed CV $\approx3\%$ (to be verified on the retrospective data) and $n=12$ per arm, $\mathrm{SE}(\hat\psi)\approx0.03\sqrt{2/12}\approx0.012$; a $4\%$ effect gives $z\approx3.3$ (power $\approx0.9$ at $\alpha=0.05$), a $2\%$ effect gives power $\approx0.4$. Hypotheses whose plausible effect is below $3\%$ are declared untestable in the prospective window and remain in the replay-only track.

\section{Registry Schema, Tool Protocol and Prompts}
\label{app:registry}

\paragraph{Tables.}
{\raggedright\texttt{hypotheses(id, text, formal\_spec, state, logit\_belief, created\_by, created\_at, parent\_ids)};
\texttt{predictions(id, hypothesis\_id, target\_ride, distribution\_json, registered\_at, scored\_at, crps, logscore, brier)};
\texttt{evidence(id, hypothesis\_id, fit\_id, log\_bf, method, recorded\_at)};
\texttt{fits(fit\_id, model\_spec\_hash, data\_cutoff, posterior\_path, loo, created\_at)};
\texttt{events(id, hypothesis\_id, kind, payload, actor, at)}.\par}

\paragraph{Invariants enforced by the store.}
(i)~\texttt{logit\_belief} is written only by \texttt{record\_evidence}; (ii)~\texttt{record\_evidence} rejects a \texttt{fit\_id} absent from \texttt{fits} or with \texttt{data\_cutoff} later than the target ride; (iii)~\texttt{predict} rejects \texttt{registered\_at} $>$ ride start; (iv)~transition to \textsc{supported} requires $\pi_H\geq0.8$, to \textsc{refuted} $\pi_H\leq0.2$, to \textsc{undetermined} a test budget exhausted with $|\log\mathrm{BF}_{\text{cum}}|<\log3$; (v)~\texttt{refine}/\texttt{merge} require a parent with a verdict or evidence; (vi)~every call is logged with its actor; (vii)~a single piece of evidence is capped at $|\log\mathrm{BF}|\leq10$; (viii)~directional evidence from a parameter posterior is admitted only if the posterior standard deviation shrank by $\geq10\%$ relative to the prior (identifiability gate); (ix)~a prediction is never deleted: if its target can no longer be observed it is marked \texttt{withdrawn} with a reason, and a prediction whose outcome exists cannot be withdrawn; (x)~an observation (\texttt{outcome\_ref}: activity and segment) scores at most one prediction. Each invariant has a unit test; \texttt{record\_evidence} additionally rejects any caller other than the statistician tool.

\paragraph{Hypothesizer prompt (abridged).}
\begin{quote}\footnotesize
\begin{verbatim}
You are a sports physiologist reviewing
diagnostics for ONE athlete. You will
receive structured residual reports from
fitted models and a list of open
hypotheses. Propose at most 3 new
hypotheses. For each:
  - claim: one sentence, specific to
    this athlete, falsifiable;
  - mechanism: 1-2 sentences with a
    literature pointer;
  - formal: a predicate over named model
    parameters OR a model-template
    change from the allowed list
    (add_covariate, add_interaction,
    add_changepoint, add_random_walk,
    split_by_condition);
  - prior_logodds: a number in [-2, 2]
    with one line of justification;
  - test: which segment(s), which
    outcome, how many traversals needed.
You cannot set beliefs, compute
evidence, or approve workouts. Do not
restate hypotheses already in the
registry; refine or merge instead.
\end{verbatim}
\end{quote}

\paragraph{Critic prompt (abridged).}
\begin{quote}\footnotesize
\begin{verbatim}
Review the proposed model-template
change. Reject if: it introduces a
parameter not separable from an
existing one given the listed
instruments; it uses descent heart
rate as a fitness signal; it
conditions on post-outcome
information; or an equivalent
hypothesis exists in lineage.
Return accept|revise|reject with
one paragraph.
\end{verbatim}
\end{quote}

\section{Workout Write-Back}
\label{app:push}
The pusher serializes $d^\star$ as a cycling workout with ordered steps, each with a duration or distance end condition and a heart-rate-zone or custom heart-rate target, creates it through the workout service and schedules it on the target date; the watch receives it after the phone application syncs, which the athlete is prompted to do. Because the endpoint is unofficial, the payload schema is validated against a workout created in the vendor's own interface and retrieved through the same API, and a file-based fallback (FIT workout copied to the device over USB) is maintained. All requests are paced and retried with exponential backoff; raw responses are archived.

\section{Synthetic Recovery Check}
\label{app:synthetic}
Each simulated athlete draws $\tau_1\sim\mathcal{U}(25,60)$, $\tau_2\sim\mathcal{U}(4,12)$, $k_2/k_1\sim\mathcal{U}(1.2,3)$, $\gamma_d\sim\mathcal{U}(1,8)$, $\gamma_{dT}\sim\mathcal{U}(0,1)$, a segment network subsampled from the real one, a ride-frequency process matched to the real athlete's calendar, heart-rate noise and dropout rates matched to measured quality statistics, and a weather series bootstrapped from the real archive. We simulate (a)~the historical load history for replay-pipeline calibration and (b)~prospective seasons under information-directed versus conventional policies. Reported: parameter-recovery coverage of $90\%$ intervals; information accrual curves; CRPS/ECE of the full pipeline on simulated held-out weeks.

\paragraph{Recovery under the athlete's actual design.}
As a first check we simulated HR-adjusted climb speeds from $\tau_1=35$, $\tau_2=7$, $k_2/k_1=2$ and a $6\%$ Student-$t$ noise on the real load history and the real $188$ observation times, segments and heart rates (a $21\%$ latent performance swing), then refit the bounded model. Over three seeds the posterior median of $\tau_1$ was $45$, $58$ and $46$ days with $90\%$ intervals $[28,65]$, $[41,76]$ and $[30,66]$ (two of three covering the truth); $\tau_2$ and $k_2/k_1$ were recovered in all three; the latent performance trajectory correlated $0.95$--$0.99$ with the truth. The design identifies \emph{performance} but not the \emph{time constants}.

\section{Data Inventory}
\label{app:data}

Activities by type: $265$ strength sessions ($274$\,h), $222$ mountain-bike rides ($295$\,h), $45$ walks, $22$ mountaineering days, $19$ treadmill runs, $15$ overland trips, $13$ trail runs and $52$ others; all $653$ have FIT files. Monthly ride counts range from $1$ (February 2026) to $22$ (August 2026), with a winter trough from December to March. One device (Forerunner~965, firmware $22.24$--$29.05$) recorded every activity. Weather records exist for every ride.

Daily wellness completeness over \NDays{} days: readiness $98\%$, acute/chronic load $98\%$, resting heart rate, stress and body battery $99\%$, sleep and overnight HRV $77\%$ (the watch was not worn on some nights). Daily load: mean exponential TRIMP $55$ per day with $116$ rest days; the vendor's proprietary load correlates with exponential TRIMP at $r=0.87$.

\end{document}